\documentclass[accepted]{uai2025} % for initial submission
\usepackage[american]{babel}
\usepackage{natbib} % has a nice set of citation styles and commands
\usepackage{mathtools} % amsmath with fixes and additions
\usepackage{booktabs} % commands to create good-looking tables
\usepackage{tikz} % nice language for creating drawings and diagrams

\usepackage{amsmath}
\usepackage{amssymb}
\usepackage{mathtools}
\usepackage{amsthm}

\usepackage{subfig}
\usepackage{booktabs} % for professional tables
\usepackage{multirow}
\usepackage[most]{tcolorbox}
\usepackage{stfloats}

\usepackage[capitalize,noabbrev]{cleveref}

\usepackage{graphicx} % Required for inserting images
\usepackage{bbm}
\usepackage{amsmath,amssymb,amsfonts}
\usepackage{appendix}
\usepackage{xcolor}

\usepackage{algorithm}
\usepackage[noend]{algorithmic}

\usepackage{ragged2e}

\theoremstyle{plain}
\newtheorem{theorem}{Theorem}[section]

\theoremstyle{definition}

\newtheorem{assumption}[theorem]{Assumption}
\theoremstyle{remark}

\def\dataset{\ensuremath{\mathcal{D}}}
\def\x{{\mathbf{x}}}

\def\b{{\mathbf{b}}}

\def\z{{\mathbf{z}}}
\def\C{\ensuremath{\mathcal{C}}}
\def\real{\mathbb{R}}

\def\dataset{\ensuremath{\mathcal{D}}}
\def\localdataset{\dataset^u}

\def\clients{\ensuremath{\mathcal{U}}}
\def\model{\theta}

\def\globallossfunction{{\loss}}
\def\locallossfunction{{\loss^u}}

\def\W{\mathbf{W}}

\def\g{\mathbf{g}}

\def\Win{\W^{(1)}}
\def\Wout{\W^{(2)}}
\def\bin{\b^{(1)}}
\def\bout{\b^{(2)}}
\def\gi{\mathbf{g}_{i}}

\def\intervals{\ensuremath{\mathcal{I}}}
\def\obs{\ensuremath{\mathcal{G}}}
\def\bini{b_i^{(1)}}
\def\obsnew{\obs_{\mathrm{new}}}
\def\loss{\ensuremath{\mathcal{L}}}

\def\batch{\ensuremath{\mathcal{B}}}

\def\prob{\mathbb{P}}

\newcommand{\eps}{\ensuremath{\varepsilon}}
\newcommand{\relu}{\textrm{ReLU}}

\title{Cutting Through Privacy:\\ A Hyperplane-Based Data Reconstruction Attack in Federated Learning}

\author[1,2]{\href{mailto:<francesco.diana@inria.fr>}{Francesco~Diana}{}}
\author[1,2,3,4]{André~Nusser}
\author[1,2,3,4]{Chuan~Xu}
\author[1,2]{Giovanni~Neglia}

\affil[1]{%
    Université Côte d'Azur \\
    France
}
\affil[2]{%
    Inria, \\
    France
}
\affil[3]{%
    CNRS\\
    France
  }
\affil[4]{%
    I3S\\
    France
  }

\begin{document}

\maketitle

\begin{abstract}

%Federated Learning (FL) enables collaborative training of machine learning models across distributed clients without sharing raw data, ostensibly preserving data privacy. However, recent studies have revealed critical vulnerabilities, showing that a malicious central server can manipulate model parameters sent to clients to reconstruct their private training data. Existing data reconstruction attacks often rely on assumptions about the clients' data distribution or degrade significantly in quality as batch sizes exceed just a few tens of samples.

%In this work, we introduce a novel data reconstruction attack that overcomes these limitations and allows for perfect recovery of the input. Our method crafts malicious model parameters to  recover arbitrarily large batches of data in classification tasks without requiring prior knowledge of client data. Through extensive experiments on image and tabular datasets, we demonstrate that our attack outperforms existing sparsity-based methods and achieves perfect reconstruction of data batches two orders of magnitude larger than the state of the art.

Federated Learning (FL) enables collaborative training of machine learning models across distributed clients without sharing raw data, ostensibly preserving data privacy. Nevertheless, recent studies have revealed critical vulnerabilities in FL, showing that a malicious central server can manipulate model updates to reconstruct clients' private training data. Existing data reconstruction attacks have important limitations: they often rely on assumptions about the clients' data distribution or their efficiency significantly degrades when batch sizes exceed just a few tens of samples.

In this work, we introduce a novel data reconstruction attack that overcomes these limitations. Our method leverages a new geometric perspective on fully connected layers to craft malicious model parameters, enabling the perfect recovery of arbitrarily large data batches in classification tasks without %requiring 
any prior knowledge of clients' data. Through extensive experiments on both image and tabular datasets, we demonstrate that our attack outperforms existing methods and achieves perfect reconstruction of data batches two orders of magnitude larger than the state of the art.

% Federated Learning (FL) enables collaborative training of machine learning models across distributed clients without sharing raw data, ostensibly preserving data privacy. However, recent studies have revealed critical vulnerabilities, showing that a malicious central server can manipulate model parameters sent to clients to reconstruct their private training data. Existing data reconstruction attacks often rely on assumptions about the clients' data distribution or degrade significantly in quality as batch sizes exceed just a few tens of samples.

% In this work, we introduce a novel data reconstruction attack that overcomes these limitations. Our method crafts malicious model parameters to recover arbitrarily large batches of data in classification tasks without requiring prior knowledge of client data. Through extensive experiments on image and tabular datasets, we demonstrate that our attack outperforms existing sparsity-based methods and achieves perfect reconstruction of data batches two orders of magnitude larger than the state of the art.

\end{abstract}

\section{Introduction}

Federated Learning (FL) enables collaborative machine learning model training across distributed clients without directly sharing data with a central server \citep{mcmahan2017communication}.
%While not sharing data aligns with the principle of data minimization \cite{bonawitz_privacy} and ostensibly enhances privacy, several attacks have revealed existing privacy vulnerabilities in FL. By inspecting the received updates and leveraging publicly available information, the central server can infer training data membership \cite{melis_memb_inf, pmlr-v139-choquette-choo21a, shokri_mia} or infer clients' sensitive attributes \cite{aia_fl}. While these attacks reveal only partial information or data-related properties, a more advanced class of data reconstruction attacks enables the server to recover entire private training data points. Such attacks have been tested in both the scenarios of a \emph{honest-but-curious} server \cite{dlg, idlg, geiping_gi, yin_gi, kariyappa23a_cocktail, dimitrov2022data, dimitrov2024spear} or under the \emph{malicious model} \cite{fowl2022robbing, curious, mkor, ZHANG2023119421, loki} threat. 
Although not sharing data ostensibly guarantees privacy, several attacks have revealed existing privacy vulnerabilities in FL. By inspecting the updates received and using publicly available information, the central server can infer training data membership \citep{melis_memb_inf, shokri_mia, pmlr-v139-choquette-choo21a} or infer clients' sensitive attributes \citep{aia_fl,diana2024attributeinferenceattacksfederated}. %hese attacks reveal only partial information or data-related properties, a more advanced 
These attacks only expose limited information, whereas more advanced 
%class of 
data reconstruction attacks enable the server to fully recover private training data points. These attacks have been %tested 
proposed in two primary settings: 
the \emph{honest-but-curious} setting, where the server passively observes updates \citep{dlg, idlg, geiping_gi, yin_gi, kariyappa23a_cocktail, dimitrov2022data, dimitrov2024spear}, and the \emph{malicious} setting \citep{fowl2022robbing, curious, ZHANG2023119421, loki, mkor, garov2024hiding}, where the server actively manipulates model parameters.
%both scenarios of a \emph{honest-but-curious} server, where the server passively observes updates \cite{dlg, idlg, geiping_gi, yin_gi, kariyappa23a_cocktail, dimitrov2022data, dimitrov2024spear}, and under the \emph{malicious model} threat \cite{fowl2022robbing, curious, ZHANG2023119421, loki, mkor}, where the server actively manipulates model parameters.
%In the first case, the attacker does not modify the model parameters, but existing works fail to recover large batch sizes. In the malicious case instead, some of the attacks partially recover batches of 256 images, but never achieved a perfect reconstruction of the full batch. Moreover, most of the literature focused on testing attacks on image datasets. While, image classification is a fundamental task in machine learning, attacks on tabular data have been highly overlooked, and few \cite{tableak, dimitrov2024spear} have proposed attacks in this domain, even if large amounts of private information, such as clinical or financial data, fall in this domain.
In the honest-but-curious setting, existing methods struggle to perfectly recover the training data for batch sizes of a few dozen 
%more than 25 
data points~\citep{dimitrov2024spear}) or 
%(e.g., more than 25 in~\citep{dimitrov2024spear}) or 
%fail to reconstruct partial representations for larger batch sizes 
produce very noisy reconstructions for a large portion of data points in batches exceeding 100 data points~\citep[Fig.~6]{kariyappa23a_cocktail}.
% and \cite{garov2024hiding}.
%(e.g., see~\citep[Fig.~6]{kariyappa23a_cocktail}).
In the malicious setting, existing  attacks~\citep{wen2022fishing, fowl2022robbing, curious, ZHANG2023119421, garov2024hiding} perform better but have not succeeded in perfectly recovering a batch of 256 data points.
%produce very noisy reconstructions for a large portion of data points in batches exceeding 100 data points~\citep[Fig.~6]{kariyappa23a_cocktail} and \cite{garov2024hiding}.
%(e.g., see~\citep[Fig.~6]{kariyappa23a_cocktail}).

%In the malicious case, some attacks \citep{wen2022fishing, curious, ZHANG2023119421} managed to partially reconstruct image batches of size 256, but none has achieved perfect recovery of an entire batch of this size.
Furthermore, most previous work has focused on image datasets, even though many real-world applications involve tabular data---such as clinical and financial records---which often contains highly sensitive information.
%despite the fact that many real-world applications involve tabular data, such as clinical and financial records, which contain highly sensitive information.

In this paper we focus on the \emph{malicious} setting, in which the server is able to craft the model parameters to improve the attack's effectiveness.
We present a novel attack that reconstructs client data by combining geometric search in the feature space with precise control over each data point's contribution to the client's updates. The core idea behind our attack is to isolate individual data points within strips defined by parallel hyperplanes and then iteratively reconstruct the entire batch by leveraging the knowledge of per-data-point gradient contributions.
% We present an attack that combines geometric searching of the client's data points in their feature space with a method to control %the gradients that enables perfect reconstruction.

% The core idea behind our attack is to identify strips within the feature space, defined by parallel hyperplanes, where each strip contains exactly one data point. By manipulating model parameters to maintain constant gradients throughout the identification process, the attacker can iteratively and systematically reconstruct all points in a batch.
% To achieve this, we first randomize but fix the (initially unknown) gradients of the loss incurred by single input points, to subsequently reconstruct the inputs in the order of separation by inductively using the reconstructed points and their associated gradients.
%
%We thereby demonstrate that tabular data, as well as any other data modality used in classification tasks, is highly vulnerable to reconstruction attacks, regardless of the batch size.
%This approach works for all classification tasks, including those involving low-dimensional tabular data, where existing state-of-the-art attacks perform particularly poorly, demonstrating its robustness across diverse data modalities.
This approach is effective for all classification tasks, including those involving low-dimensional tabular data where existing state-of-the-art attacks perform poorly, demonstrating its robustness across diverse data modalities.
Concretely, our key contributions are the following:
\begin{itemize}[itemsep=0pt,topsep=0pt]
    \item We derive an analytical upper bound on the accuracy of existing data reconstruction attacks on random data.
    % \item We demonstrate that, in the presence of a fully connected layer, a malicious central server can control the gradients to amplify the privacy leakage from clients' updates.
    \item We show how a malicious server can control each data point's contribution to the client's updates.
    %We demonstrate that a malicious central server  can control the %gradients of a fully connected (FC) neural network, 
    %model weights
    %significantly amplifies privacy leakage of clients' updates.
    \item We propose a perfect reconstruction algorithm for classification tasks that is agnostic to input dimensionality. Hence, while most of the existing literature primarily addresses image classification, our attack can reconstruct both image and tabular data.
    \item Our method enables full recovery of all samples within a batch and remains effective regardless of batch size, provided that the number of training rounds is sufficiently high.
    \item We validate our findings on both image and tabular datasets, achieving perfect reconstruction on large batches of up to 4,096 data points, where %existing methods fail to recover batches of 64 inputs. 
    our baseline fails to recover batches of 64 inputs.
\end{itemize}

%DRAFT
% \begin{enumerate}
%     \item The goal of the existing methods is to isolate points on the convex hull.
%     \begin{itemize}
%         \item The performance of these methods is upper bound by the number of points on the convex hull. For the low-dimensional case, the attack accuracy drops to close to 0 as the batch size grows to infinity (see Theorem 1).
%         \item For high-dimensional case, where all the existing methods tested on, the attack performance is highly influenced by the batch size. They can work up to 200 images...
%     \end{itemize}
%     \item Our contribution
%     \begin{itemize}
%         \item Provide a perfect reconstruction algorithm for all-dimensional inputs on (classification) problem, which can reconstruct all the samples in the batch and insensitive to the batch size as long as the number of the attack rounds are sufficient.
%     \end{itemize}
% \end{enumerate}

\section{Background}
\label{sec:background}

In federated learning, a set of users $\clients$, orchestrated by a central server, cooperate to optimize a shared model $\theta$, which minimizes the weighted sum of clients' empirical risks: 
\begin{equation}\label{eq:globalobjective}
    \min_{\model \in \real^d} \globallossfunction(\model) = \sum_{u\in \clients} p^u\loss(\model, \localdataset),
\end{equation}
where $\localdataset = \{(\x_i^u, y_i^u)\}_{i=1}^{n^u}$ is the local dataset of  client~$u$, $\locallossfunction$ measures the loss of  model $\theta$ on points in $\localdataset$,  and $p^u$ is a positive weight quantifying the relative importance of  client $u$, s.t.\ $\sum_{u\in \clients} p^u = 1$. Typical choices in FL are $p^u = 1/|\clients|$ and $p^u \propto n^u$.

Federated Learning typically involves iterative algorithms that operate in multiple communication rounds between clients and the central server. 
At each communication round, clients receive the current global model from the server and transmit updates computed on their local datasets. The nature of the updates depends on the specific federated optimization algorithm in use. For instance, in FedSGD \citep{45187, mcmahan2017communication}, each client computes and transmits a single gradient update $\nabla_{\theta}\loss(\theta, \localdataset)$. In contrast, algorithms like FedAvg \citep{mcmahan2017communication} allow clients to perform multiple local stochastic gradient descent steps before sending their updated model parameters to the server.

While FL preserves data locality, the transmitted updates---whether gradients or model parameters---can 
leak 
%encode 
sensitive information about the clients' datasets \citep{melis_memb_inf,diana2024attributeinferenceattacksfederated,fowl2022robbing}. 
%This makes FL inherently vulnerable to 
This vulnerability  exposes FL to adversarial attacks 
%, particularly those 
aimed at reconstructing private data \citep{dlg,fowl2022robbing,curious,dimitrov2024spear}. The success rate of such attacks depends on the adversary's capabilities.
In what follows, we consider the attacker to be the server.
%In such attacks, an adversary attempts to reconstruct clients' local data by analyzing the updates exchanged during training, with a success rate that inevitably depends on his capabilities. 

Two adversarial settings are commonly considered in the literature: the \emph{honest-but-curious} setting \citep{dlg, dimitrov2024spear} and the \emph{malicious} setting \citep{fowl2022robbing, curious}.
In the honest-but-curious setting, the server passively observes updates without interfering with the training process \citep{Paverd2014ModellingAA}.
In the malicious setting, the server actively manipulates the training process by transmitting modified models designed to induce privacy leakage. %\citep{fowl2022robbing, curious}.
In both settings, two main categories of attacks have been proposed: optimization-based attacks and analytical attacks.

\subsection{Optimization-based attacks}
%Optimization-based attacks treat the reconstruction problem as an optimization task. 
In optimization-based attacks, the server attempts to reconstruct the client's data by minimizing the difference between the actual update sent by the client to the server (e.g., the gradient or updated model) and the update that would have been computed by the client from the reconstructed data. These attacks often begin with dummy inputs, which are progressively refined through iterative steps.

%, aiming to recover private data by starting with a dummy input and iteratively refining it to minimize the distance between the client's actual update (e.g., the gradient) and the update would have been computed starting from the estimated input.
%In these attacks, the server iteratively estimates the client's private data as follows: 1) it computes the update that the client would have sent to the server based on its current estimation of the private data, and 2) it refines the estimate to minimize the difference between the computed update and the actual observed update.

%true gradient and the pseudogradient computed from the dummy input. 
% Early works \citep{dlg, idlg, geiping_gi, yin_gi} primarily targeted the recovery of image inputs %from small batch 
% sizes within the honest-but-curious setting, but were successful only on small batch sizes. To overcome the batch size limitation and enable recovery with secure aggregation in place \citep{sec_agg}, \cite{kariyappa23a_cocktail} reformulated the reconstruction task as a blind source separation problem, successfully partially extracting images from large batches. \cite{dimitrov2022data} designed an optimization attack capable of recovering images from a single client trained using FedAvg, and \cite{tableak} introduced the first optimization-based attack targeting tabular data, achieving partial recovery of the input.

Early works \citep{dlg, idlg, geiping_gi, yin_gi, dimitrov2022data} proposed the recovery of image inputs %from small batch 
within the honest-but-curious setting, but were successful only on small ($\le 50$) batch sizes and, even then, produced noisy reconstructions. 
%To extend the scope of these attacks and  
%of the attack the batch size limitation and 
%enable recovery in presence of secure aggregation~\citep{sec_agg}, % in place, 
\cite{kariyappa23a_cocktail} reformulated the reconstruction task as a blind source separation problem, enabling recovery in presence of secure aggregation~\citep{sec_agg}, however, they still only obtained very noisy images on batches larger than 200 data points (see Figure~6 in their work).
%extract noisy approximations of a fraction of images from large batches. 

% \citet{garov2024hiding} tackled the same problem by training an encoder-decoder network, though requiring auxiliary data.  While their approach produces high-quality images, it successfully reconstructs only half of the images in a batch of 512.
%\cite{dimitrov2022data} designed an optimization attack capable of recovering images from a single client trained using FedAvg, and 
\cite{tableak} introduced the first optimization-based attack targeting tabular data and reported a 70\% reconstruction success rate on a batch of 128~samples.
%, but failing to perfectly recover of the input. 

Overall, these attacks struggle with the challenge posed by gradient aggregation, and do not scale effectively to large batch sizes without requiring auxiliary data.
% To address this limitation, more recent works have started to consider a malicious server that can improve his capabilities by modifying model's parameters and architecture.
To address these shortcomings, alternative approaches consider a scenario where the server is malicious.
In this context, \citet{wen2022fishing} proposed a \emph{fishing} strategy that alters the weights of the classification layer of a neural network to amplify the gradient contribution of a single input within a batch, increasing the success rate of the optimization step. 
%By repeating this process sequentially over multiple communication rounds, the attacker can disaggregate the gradient contributions of inputs belonging to a specific class within the client batch, thereby improving the accuracy of subsequent optimization steps.
However, their approach relies on estimating the distribution of one input feature, which the authors assume to follow a normal distribution. While this assumption may hold true for image data, it is often unrealistic for tabular datasets, where features can be binary or categorical.
Alternatively, \citet{garov2024hiding} designed an encoder-decoder architecture to isolate individual input gradients from the aggregated update in a latent space using an encoder to reduce the detectability, and subsequently reconstruct the original input with a decoder. %In their approach, the attacker optimizes the parameters sent to each client to facilitate the recovery of a specific example from a batch. 
Nonetheless, their method depends on access to auxiliary data for training the encoder-decoder network and struggles to achieve high-quality recovery for approximately half of the images in a batch of 512 samples.

\subsection{Analytical attacks}
Unlike optimization-based attacks, analytical methods aim to exactly recover input data by exploiting the properties of fully connected layers in neural networks. Prior works \citep{phong, geiping_gi} have shown that the input to a biased linear layer followed by a ReLU activation function can be exactly reconstructed from its gradients. 
In what follows, we describe this technique, which forms the core of our attack.

%This property enables perfect input recovery when this layer serves as the network’s input layer.
For simplicity, consider a client training a model where the first layer is a fully connected (FC) layer with input vector $\x \in \mathbb{R}^{d}$ and output vector $\z \in \mathbb{R}^{N}$. Let $\W \in \mathbb{R}^{N \times d}$ and $\b \in \mathbb{R}^{N}$ denote the weight matrix and the bias vector, respectively. 
Let the output of the $i$-th neuron in the layer be $z_i = \relu(\W_i \x + b_i)$. 
Given a sample $(\x_j, y_j)$, let $\loss_j$ and $z_{i,j}$ denote the loss and the output of neuron~$i$ for this sample, respectively. The gradient of the loss with respect to $\W_i$ and $b_i$ can be computed as follows:\footnote{
 The expression $\frac{\partial\mathbf{f}}{\partial \x}$, where $\mathbf{f}: \mathbb R^d \to \mathbb{R}^N$, can also be interpreted as the Jacobian of the function $\mathbf f$, with the (less common) convention that 
$\left(\frac{\partial\mathbf{f}}{\partial \x}\right)_{i,j} = \frac{\partial f_j}{\partial x_i}$.
}
$\frac{\partial \loss_j}{\partial \W_i} =  \frac{\partial \loss_j}{\partial z_i} \frac{\partial z_i}{\partial \W_i} = \frac{\partial \loss_j}{\partial z_i} \x_j \mathbbm{1}_{z_{i,j}>0}$ and $\frac{\partial \loss_j}{\partial b_i} = \frac{\partial \loss_j}{\partial z_i}   \frac{\partial z_i}{\partial b_i}= \frac{\partial \loss_j}{\partial z_i}\mathbbm{1}_{z_{i,j}>0}$, from which it follows that $\frac{\partial \loss_j}{\partial \W_i} =   \frac{\partial \loss_j}{\partial b_i} \x_j $.
Consequently, as long as $\frac{\partial \loss_j}{\partial b_i} \neq 0$ (which requires that $\x_j$ activates neuron $i$, i.e., $\W_i \x_j + b_i > 0$), knowing the gradient of the model loss enables exact reconstruction of the input:
%~\citep{phong, geiping_gi}:
\begin{equation}\label{eq:single_input_rec}
    \x_j=\frac{\partial \loss_j}{\partial \W_i} \left(\frac{\partial \loss_j}{\partial b_i}\right)^{-1}.
\end{equation}
However, in practice, gradients are computed on a batch $\batch=\{(\x_j, y_j)\}_{j=1}^n$ of size $n$, rather than on a single input. %Let $\loss_j$ denote the loss of the model on sample $(\x_j, y_j)$. 
The attacker can then only observe the  gradient of the average loss $\loss = \frac{1}{n} \sum_{j=1}^n \loss_j$:
\begin{equation}\label{eq:aggr_obs}
     \frac{\partial \loss}{\partial \W_i} =  \frac{1}{n} \sum_{j=1}^n \frac{\partial \loss_j}{\partial b_i}\x_j \, \\
\end{equation}
\begin{equation}\label{eq:aggr_coeff}
     \frac{\partial \loss}{\partial b_i} =  \frac{1}{n} \sum_{j=1 }^n \frac{\partial \loss_j}{\partial b_i}.\\
\end{equation}
If $\frac{\partial \loss}{\partial b_i}\neq 0$---implying that at least one input activates neuron~$i$---the attacker can reconstruct a combination $\g_i$ of the inputs from this gradient \citep[Theorem~1]{ZHANG2023119421}:
%(i.e., at least one input activates neuron $i$), 

\begin{align}
    \g_i &= \frac{\partial{\loss}}{\partial{\W_i}} \cdot \left( \frac{\partial{\mathcal{L}}}{\partial{b_i}} \right) ^{-1} = \sum_{j=1}^n \alpha_j \x_j, \label{eq:obs_derivation} \\
    \alpha_j &=  \frac{\frac{\partial{\loss_j}}{\partial{b_{i}}}} {\sum^{n}_{k=1} \frac{\partial{\loss_k}}{\partial{b_{i}}}}.\label{eq:alpha_derivation}
\end{align}
Equations~\ref{eq:obs_derivation}  and \ref{eq:alpha_derivation} are the starting points for different attacks.

\subsubsection{Sparsity-Based Attacks}
%Although batch aggregation only allows for the recovery of linear combinations of inputs, the inherent data-leaking property of ReLU layers remains appealing for attackers.
Several works, commonly referred to as \textit{sparsity-based} attacks, exploit the sparseness induced by the ReLU activation to recover individual data points.
In the honest-but-curious setting, \cite{dimitrov2024spear} proposed a method that leverages the low-rank structure of gradients and exploits ReLU-induced sparsity to efficiently guide a greedy search that enables the recovery of the inputs.
% Their attack achieves perfect recovery for batches with at most $25$ data points, but fails to scale to larger batch sizes due to exponential computational cost. 
Their attack achieves perfect recovery for batches with $\leq 25$ data points but fails to scale to larger batch sizes due to the exponential computational cost w.r.t.~the number of samples.

Alternatively, some approaches introduced malicious model modifications to enforce sparsity in the gradients and increase the percentage of single inputs that activate a neuron. \cite{curious} presented a \emph{trap weights} method, where the  FC layer's weights are randomly initialized, with one half having  positive values and the other half larger negative values. The idea behind this approach is to maximize the probability that a single input activates a given neuron and can be reconstructed using~\eqref{eq:single_input_rec}. 
%minimize the number of inputs that activate a neuron, enabling perfect input reconstruction. 
%Building on the same intuition, 
\cite{ZHANG2023119421} extended this attack to networks with sigmoid activation, though their method requires additional data to fine-tune the malicious weights effectively. \cite{pasquini_sec_agg} further leveraged ReLU-induced sparsity to bypass secure aggregation \citep{sec_agg}.

As we will demonstrate in Sec.~\ref{sec:limit}, sparsity-based attacks are significantly less effective on low-dimensional data, such as tabular datasets.
%the effectiveness of the sparsity-based attacks on low-dimensional data, such as  tabular datasets, is limited.

% \cite{curious} presented a \emph{trap weights} method, where the FC layer's weights are randomly initialized, one half with positive weights and the other half with negative weights of large magnitude. The idea behind this approach is to minimize the number of inputs that activate a neuron, enabling perfect input reconstruction. Building on the same intuition, \cite{ZHANG2023119421} extended the attack to networks with sigmoid activation, though their method requires additional data to fine-tune the malicious weights effectively. \cite{pasquini_sec_agg} further leveraged ReLU-induced sparsity to bypass secure aggregation \citep{sec_agg}.
% However, as we demonstrate in Section~\ref{sec:limit}, the effectiveness of these attacks on low-dimensional data, such as tabular datasets, is inherently limited.

\subsubsection{Other Analytical Attacks}
Some analytical attacks can recover input data without inducing gradient sparsity.
\cite{fowl2022robbing} assume prior knowledge of the distribution of a linear combination of client inputs and configure neuron biases such that the gradients with respect to different neurons’ weights differ by only a single input, which can be recovered using~\eqref{eq:aggr_obs} and~\eqref{eq:aggr_coeff}.
%introduced a "binning" strategy to disaggregate gradients contributions of individual images within a batch.
%Their method 
%assume knowledge of the distribution of a linear combination of the inputs in the local dataset and use this information to select bias values for all neurons, ensuring that multiple combinations in~\eqref{eq:aggr_obs} and \eqref{eq:aggr_coeff} differ by a single input, which can then be easily decoded.
%Their method assumes to know the distribution of a linear combination of the inputs $\mathbf{w}^T \x$ over the local dataset and uses this information to select bias values for all neurons in order to probabilistically guarantee that multiple combinations in~\eqref{eq:aggr_obs}
%and \eqref{eq:aggr_coeff} differ by a single input that can be easily decoded. 
%%%% OLD
% Our parallel search method in Sec.~\ref{sec:parallel} is similar to their approach but does not require prior knowledge of a distribution, as we dynamically discover the appropriate biases. Additionally, our attack can span multiple communication rounds.
%%%% NEW
The first round of our parallel search method presented in Sec.~\ref{sec:parallel} is similar to their approach but there are two crucial differences: (i) they require prior knowledge of a distribution, while we dynamically discover the appropriate biases, and (ii) they do not propose any iterative procedure that enables a bias search across multiple communication rounds, as we do.
%they do not present means to retain relevant information across multiple communication rounds, ruling out a search approach (like ours).
%While their method perfectly recovers fractions of large batches, it relies on prior knowledge of client data and strong assumptions about its distribution. Specifically, they assume that it is possible to estimate a continuous quantity from the data, which they presume follows a Gaussian or Laplace distribution. However, this assumption is unrealistic for tabular data, where features may be categorical or strictly binary, making such estimates unreliable.
% \francesco{Initially, for the robbing attack I wanted also to criticize that their attack cannot be applied for multiple rounds, but I am not sure that I can. What they can do for example is that, after the first round, they can place the bins to "disaggregate" the bins that were activated by multiple images at the previous round, potentially recovering the full batch}.

\cite{loki} adapted this method to break secure aggregation, designing convolutional filters to isolate individual client's gradient contributions.

\subsubsection{Extension to other architectures}
\label{sec:extensions}
% All the analytical attacks discussed so far rely on the fully connected attack layer being the first layer of the network or having only fully connected layers preceding it for successful data recovery. Attacks specifically targeting convolutional neural networks (CNNs) have also been explored. \cite{mkor} proposed specific parameters' modifications to enable analytical recovery from CNNs. Additionally, \cite{curious} proposed to adapt analytical attacks to CNNs by adjusting convolutional layer weights to behave as identity mappings. By doing so, the attacker can recover the original input from the first fully connected layer in the network. However, pooling layers and dropout can mitigate the attack by disrupting the gradient structure and reducing search accuracy.

% An alternative approach, proposed by \citet{kariyappa23a_cocktail} and \citet{ZHANG2023119421}, is to reconstuct the feature maps that serve as inputs to the first fully connected layer. Once these feature maps are obtained, the attacker can solve an optimization problem to recover the original input . 
%Until now, we discussed the reconstruction attacks when the attacked layer is the first layer in the network. The attacks extend immediately if this layer is preceded by other FC layers. 
%All the analytical attacks discussed so far rely on the fully connected attack layer either being the first layer of the network or being preceded only by fully connected layers, for successful data recovery.
Until now, we have discussed reconstruction attacks targeting the first layer in the network. These attacks can be directly extended when the attacked layer is preceded by additional FC layers.

Several works have explored adapting these attacks to convolutional neural networks (CNNs). \cite{curious} proposed modifying the weights of convolutional layers to act as identity mappings, allowing the attacker to recover the original input from the reconstructed input of the first FC layer. However, pooling layers and dropout can disrupt the gradient structure, reducing the effectiveness of the recovery process. An alternative approach, introduced by \citet{kariyappa23a_cocktail} and \citet{ZHANG2023119421}, focuses on reconstructing the feature maps that serve as inputs to the first fully connected layer. Once these feature maps are obtained, the attacker solves an optimization problem to recover the original input. Additionally, \cite{mkor} proposed specific parameter modifications to facilitate analytical recovery in CNNs, further broadening the applicability of these attacks.

Other works have specifically targeted transformer-based models. \citet{lu2022april} demonstrated that analytical image recovery is feasible even in attention-based networks; however, their attack is restricted to a batch size of one. \citet{fowldecepticons} and \citet{chupanning} extended their earlier work \citep{fowl2022robbing} to extract textual data from clients; however, their methods do not extend to other data modalities, such as images.
%; however, their methods remain limited to text
%and do not address other data modalities.}
% Some works explored adapting these attacks to convolutional neural networks (CNNs). \cite{curious} proposed modifying convolutional layer weights to act as identity mappings, allowing the attacker to recover the original input from the first fully connected layer in the network. However, pooling layers and dropout can disrupt the gradient structure, making recovery less effective. An alternative approach, introduced by \citet{kariyappa23a_cocktail} and \citet{ZHANG2023119421}, focuses on reconstructing the feature maps that serve as inputs to the first fully connected layer. Once these feature maps are obtained, the attacker solves an optimization problem to recover the original input. Finally, \cite{mkor} proposed specific parameter modifications to enable analytical recovery in CNNs, further extending the applicability of these attacks.

% Finally, \cite{lu2022april} demonstrated that analytical image recovery is also possible in attention-based networks.

\section{Limitations of Existing Sparsity-Based Attacks}\label{sec:limit}

The sparsity-based attacks proposed by \citet{curious, ZHANG2023119421} attempt to modify the parameters of the FC layer so that each neuron in the layer is activated only by a single input. 
Given the batch  $\mathcal{B}=\{(\x_1, y_1) \dots, (\x_n, y_n)\}$, neuron $i$ is activated only from input $\x_j$ if $\W_i \x_j+b_i>0$ and $\W_i \x_k+b_i<0$ for $k \in \{1, \dots, n\}\setminus \{j\}$.
In this section we provide a geometric interpretation of the problem and demonstrate that the accuracy of sparsity-based attacks is fundamentally constrained by the dimensionality of the input they aim to reconstruct.

%Given a batch $\mathcal{B}$ with inputs $\x$ of dimensionality $d$, 
Consider the set of inputs for which the activation value of neuron $i$ is equal to $0$, i.e., $    \{ \x \in \mathbb{R}^d \mid \W_i\x + b_i= 0\}$.
This set is a hyperplane in the Euclidean space $\mathbb R^d$ orthogonal to the vector $\W_i$.
%where $\W_i \in \mathbb{R}^d$ is the normal vector of the hyperplane and $b_i \in \mathbb{R}$ is a scalar. %Note that $-\frac{b_i}{||\W_i||}$ determines the offset of the hyperplane from the origin. 
% Let $\W_i$ and $b_i$ be the weights and the bias term for one neuron. If the neuron is activated by only one input $\x$, then we have $\W_i\x + b_i > 0$ and $\W_i\mathbf{\x'} + b_i\leq 0, \forall \x'\in \batch \backslash \{\x\}$. The subsequent ReLU activation enables the perfect separation of $\x$ as all the other samples' contributions will be 0.
The attacks proposed by \cite{curious, ZHANG2023119421} can be interpreted as identifying hyperplanes orthogonal to random directions $\W_i$ that separate a single input from the rest of the inputs in the batch, which we denote by $\batch_{\x}= \{\x_1, \x_2, \dots, \x_n\}$. Given a gradient computed on the batch $\batch$, the maximum number of recoverable points is upper-bounded by the number of points in  $\batch_{\x}$ that are linearly separable from all the others, i.e., the number of vertices on the convex hull of $\batch_{\x}$. By bounding the expected number of vertices, we derive the following theorem (proof in App.~\ref{sec:proof_convex_hull}):
%The following Theorem (proof in App.~\ref{sec:proof_convex_hull}) provides asymptotics for the expected number of vertices for different distributions.
%Hence, we can upper bound the expected expected success rate of the attacks by the expected size of the convex hull for different distributions.

\begin{theorem}
\label{theo:active_bound}
Let the input features of the client's local dataset consist of $n$ random points in $\mathbb{R}^d$ that are drawn
\begin{enumerate}
    \item uniformly at random from the unit ball, or
    \item uniformly at random from the unit hypercube, or
    \item from a centered normal distribution with covariance matrix $\mathbf{I}_d$.
\end{enumerate}
Consider training a machine learning model through FedSGD with full-batch updates.
%a classification task through FedSGD with full-batch updates.
% Any reconstruction attack that relies on isolating single inputs from the others.
% The expected number of samples that are reconstructed by using \eqref{eq:single_input_rec} in each of the above cases is upper bounded by
The expected number of samples that can be reconstructed by any attack relying on isolating individual inputs from the rest of the batch is:
\begin{enumerate}
    \item $O(n^{(d-1)/(d+1)})$, %or
    \item $O(\log^{d-1} n)$, %or
    \item $O(\log^{(d-1)/2} n)$,
\end{enumerate}
respectively for the three distributions above.
\end{theorem}

% Theorem~\ref{theo:active_bound} indicates that, for a fixed input dimension $d$, the expected fraction of reconstructed samples asymptotically approaches zero as the batch size $n$ increases, motivating the degradation in the performance of existing attacks observed by \cite{curious, ZHANG2023119421} when increasing batch sizes.
Theorem~\ref{theo:active_bound} shows that, for a fixed input dimension $d$, the expected fraction of reconstructed samples---i.e., the attack success rate---approaches zero asymptotically as the batch size $n$ increases. This result explains the attack performance degradation observed by \cite{curious, ZHANG2023119421} when larger batch sizes are used.
Furthermore, the theorem highlights a fundamental limitation in low-dimensional settings: the smaller the input dimension~$d$, the lower the success rate. 
This indicates that while such attacks can succeed in high-dimensional regimes, their effectiveness diminishes substantially when applied to low-dimensional data.
%This suggests that while these attacks may be effective for high-dimensional data, their applicability to lower-dimensional data is significantly constrained.
%Moreover,  it highlights a fundamental limitation in low-dimensional settings---the smaller the input dimension $d$ the smaller the success rate---suggesting that while these attacks may be effective for high-dimensional data, their applicability to lower-dimensional data is  constrained. 
Moreover, high-dimensional data often lies on manifolds of much lower intrinsic dimensionality~\citep{NIPS2010_8a1e808b, pope2021the, brown2023verifying}, potentially reducing the success rate of these attacks even further than predicted by Theorem~\ref{theo:active_bound}.

\section{Our Attack} \label{sec:our_attack}
% We now describe our attack.
In this section, we first present our threat model (Sec.~\ref{sec:threat}) and then introduce our attack assuming the attacker initially exploits a single neuron in the linear layer (Sec.~\ref{sec:loc_and_rec} and~\ref{sec:control_weights}). We subsequently demonstrate how the attacker can leverage all neurons in parallel to significantly accelerate the reconstruction process (Sec.~\ref{sec:parallel}).

\subsection{Threat Model}
\label{sec:threat}
In this work, we consider a malicious server whose intent is to recover clients' private data by altering model parameters, consistent with the setting described in \cite{wen2022fishing,curious,ZHANG2023119421}, where clients trust the server and have no control over the model parameters.
We note that a third party capable of intercepting communications between the client and server could also act as an attacker.

Additionally, since the central server orchestrates the FL protocol, we assume that it is responsible for selecting the clients participating in each training round. %, enabling it to target specific users. 
Unlike previous works \citep{fowl2022robbing,wen2022fishing,curious,ZHANG2023119421,loki}, we assume the server has no knowledge of the clients' data distribution and does not possess any auxiliary training data. The information available to the server is limited to some bounds on the range of the data after preprocessing.
%Following the \textit{cross-silo} setting of \cite{wen2022fishing}, 
%We consider a cross-silo FL setting and 
We suppose that clients follow the FedSGD protocol and at each round compute a full-batch gradient update. 
%Furthermore, we consider models containing a ReLU-based FC input layer and a FC classification layer. The attack can be extended to other architectures as described in other papers (see Sec.~\ref{sec:extensions}).

% In Section~\ref{sec:ex_attacks} we described how fully-connected layer followed by ReLU activation functions naturally leak data points. Here, we provide a novel method to control the loss function in classification tasks, and perfectly reconstruct a training batch of arbitrary size. In particular, we demonstrate that by fixing the weight direction in a fully-connected neural network and modifying the bias terms, it is possible to recover the individual sample contributions to the aggregated gradients $\frac{\partial{\loss}}{\partial{\W^{(1)}_i}}$ and $\frac{\partial{\loss}}{\partial{b^{(1)}_i}}$, thereby enabling the sequential reconstruction of each input in the batch.
\subsection{Locate and Reconstruct the Inputs}\label{sec:loc_and_rec}
Consider that the client under attack has dataset $\batch$ with corresponding inputs $\batch_{\x}$.
% Our attack requires the following distinguishibility assumption.
% \begin{assumption}
%     \label{ass:distinguishbility}
% There exists a distribution $\mathcal P$ over $\mathbb R^d$ such that $\textrm{Prob}_{\w \sim \mathcal P}(\exists  \x_1, \x_2 \in \batch_{\x} : |\mathbf w^T (\x_1| ) \le \delta$
% \end{assumption}
For simplicity, 
%and without loss of generality, 
we assume that the neural network consists solely of a FC input layer with ReLU activations (with parameters $\Win \in \mathbb{R}^{N \times d}$ and $\bin \in \mathbb{R}^{N}$) and a FC classification layer (with parameters $\W^{(2)} \in \mathbb{R}^{C\times N}$ and $\b^{(2)} \in \mathbb{R}^{C}$). 
%$We denote the parameters of the first layer by $\Win \in \mathbb{R}^{N \times d}$ and $\bin \in \mathbb{R}^{N}$, and the parameters of the second layer by $\W^{(2)} \in \mathbb{R}^{N\times C}$ and $\b^{(2)} \in \mathbb{R}^{C}$.
%a fully-connected neural network with two layers and a ReLU activation function, where $\Win \in \mathbb{R}^{N\times d}$, and $\bin \in \mathbb{R}^{N}$ denote the parameters of the first layer, and $\W^{(2)} \in \mathbb{R}^{N}$, and $\b^{(2)} \in \mathbb{R}^{C}$, indicate the parameters of the classification layer. 
The output of the network can be expressed as $\z^{(2)}=\W^{(2)}\z^{(1)}+\b^{(2)} \in \mathbb{R}^C$, where $\z^{(1)}~=\relu(\W^{(1)}\x+\b^{(1)}) \in \mathbb{R}^N$.
The attack can be extended to other architectures as described in other papers (see discussion in  Sec.~\ref{sec:extensions}).
%The extension of this approach to a neural network with an arbitrary number of layers is described in App.~\ref{app:gener_nn}.
%As discussed in Section~\ref{sec:limit}, $\Win_i$ can be interpreted as the normal vector of a hyperplane in the Euclidean space and $-\frac{\bini}{||\Win_i||}$ as the offset of the hyperplane from the origin.

%To facilitate the exposition, we now make an assumption here and only later, in Section~\ref{sec:control_weights}, we show how to satisfy it.

\begin{figure}[t]
    \centering
    \includegraphics[width=0.9\linewidth]{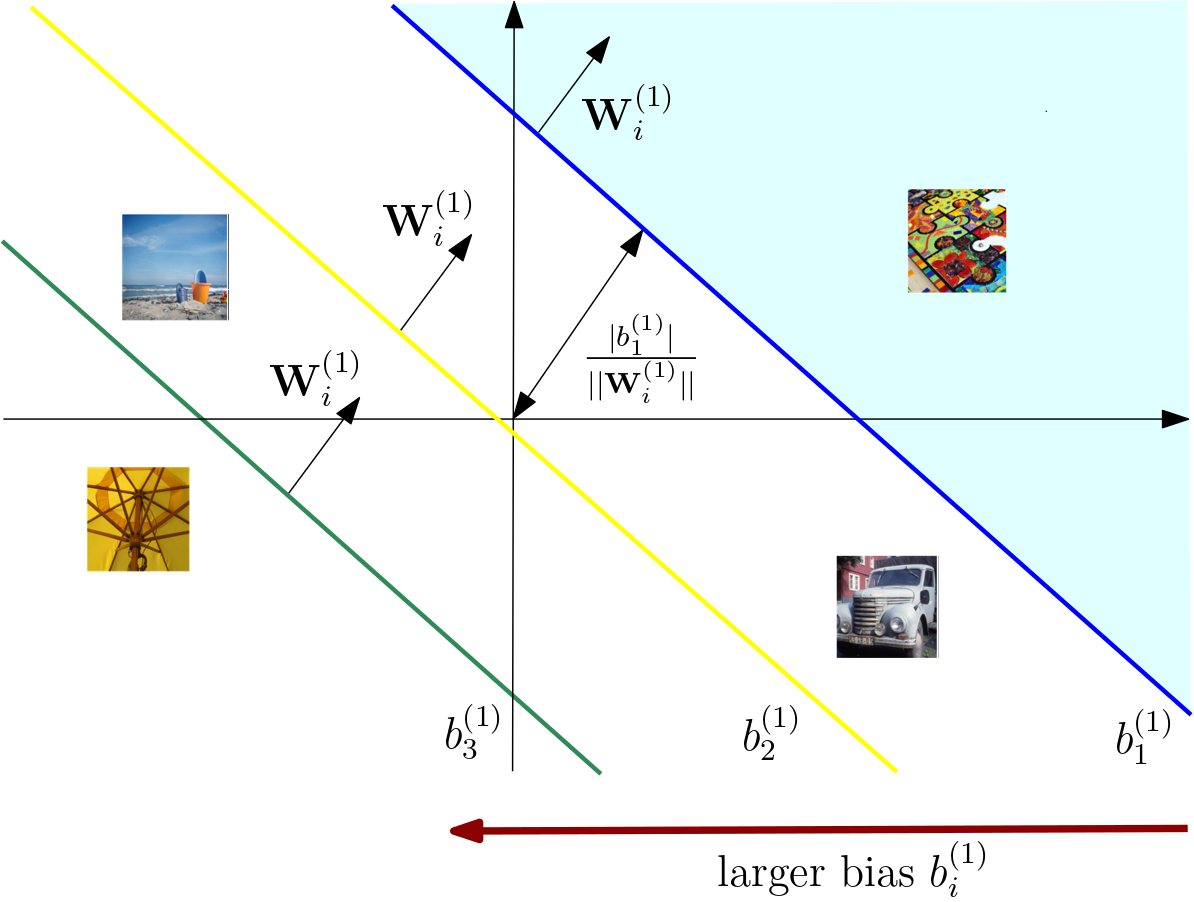}
    \caption{Illustration of our attack strategy, in which hyperplanes are positioned to isolate a single image within the strip they define. The light-blue region represents  the half-space where $\Win_i \x + b_1^{(1)} > 0$.}
    \label{fig:geometry}
\end{figure}

The server selects a random hyperplane by sampling $\Win_i$ from an arbitrary random distribution. During different communication rounds, it sends the client under attack a model whose parameters are constant except for the bias value $\bini$.
Geometrically, changing $\bini$ corresponds to translating the hyperplane through the input space (see Fig.~\ref{fig:geometry}). The goal is to find a sequence of scalars $\hat b^{(1)}_{i,0} \leq \dots \leq \hat b^{(1)}_{i,n}$ such that for each consecutive pair of hyperplanes $\Win_i \x + \hat b^{(1)}_{i,k} = 0$ and $\Win_i \x + \hat b^{(1)}_{i,k+1} = 0$, exactly one input $\x_j$ in the client's local dataset lies in-between them.
%, that is $\Win_i \x_j + b^{(1)}_{i,k} < 0$ 
Note that as the bias transitions between consecutive values in the sequence, a single additional input activates the neuron $i$.

We now explain how the server determines this sequence of bias values and how it can reconstruct the isolated samples. To facilitate the exposition, we introduce an assumption here, which we justify in Sec.~\ref{sec:control_weights}.

\begin{assumption}\label{ass:grad}
%For every input $\x_j \in \batch_{\x}$ and the values of $b^{(1)}_i$ tested by the attacker, if neuron $i$ is activated by input $\x_j$ for the specific value of $b^{(1)}_i$, the derivative $\frac{\partial \loss_j}{\partial z^{(1)}_i}$ does not depend on $b_i^{(1)}$.
For each input $(\x_j, y_j) \in \batch$ and each value of $b^{(1)}_i$
%values $b^{(1)}_i$ 
tested by the attacker, if neuron $i$ is activated by the input $\x_j$ for a given value of $b^{(1)}_i$, the derivative ${\partial \loss_j}/{\partial b^{(1)}_i}$ remains independent of $b^{(1)}_i$. %\textcolor{blue}{SUGGEST TO REMOVE: and $\alpha_{j}$~\eqref{eq:alpha_derivation} is non-negligible.}
% If an input $\x_j$ activates the $i$-th neuron, the gradient of its loss with respect to $\z^{(1)}_i$ remains constant across different communication rounds, i.e., $\forall \x_j \in \batch_{\x}$:
% \begin{equation*}
%     \frac{\partial \loss_j}{\partial z_i}(t)=\frac{\partial \loss_j}{\partial z_i}(t+1)
% \end{equation*}
% where $t \in \{1,\dots T\}$ indicates the communication round.
\end{assumption}

According to \eqref{eq:obs_derivation}, for a given bias value $b^{(1)}_{i,k}$, the server observes the client's aggregate gradient $\g_k = \sum_{j=1}^n \alpha_{j,k} \x_j$ 
% \textcolor{blue}{AS ABOVE, I WOULD REMOVE: with a non-negligible $\alpha_{j,k}$ for activated input $\x_j$ as in \eqref{eq:alpha_derivation}}. 
Given hyperplanes $\Win_i \x + b^{(1)}_{i,k} = 0$ and $\Win_i \x + b^{(1)}_{i,k+1} = 0$, the attacker can determine whether they enclose at least one point by comparing $\g_{k}$ and $\g_{k+1}$. Indeed, if there is no input point between the two hyperplanes, the same set of inputs activates neuron~$i$ and $\alpha_{i,k} = \alpha_{i,k+1}$ (because of Assumption~\ref{ass:grad}), then $\g_k=\g_{k+1}$. If $\g_k\neq\g_{k+1}$, then the server knows that there is at least one input in-between and, by binary search, can progressively isolate the inputs as far as their projections along the direction $(\W_i^{(1)})^\intercal$ differ by more than a threshold~$\varepsilon$ (see App.~\ref{app:bound_on_rounds} for its configuration rule).

After isolating all the input points, %the attacker proceeds with their reconstruction as follows. 
%We relabel the inputs so that $\x_j \in \batch_{\x}$ is the input between the two hyperplanes corresponding to $\hat b^{(1)}_{i,j-1}$ and $\hat b^{(1)}_{i,j}$. 
% The attacker
the attacker can reconstruct the inputs sequentially from $\x_1$ to $\x_n$, where we relabel the inputs so that $\x_j \in \batch_{\x}$ is the input between the two hyperplanes corresponding to $\hat b^{(1)}_{i,j-1}$ and $\hat b^{(1)}_{i,j}$. 
Input $\x_1$ can be directly reconstructed using \eqref{eq:single_input_rec} with the client's loss gradient computed for $\hat b^{(1)}_{i,1}$.
%and $\alpha_{}$.
We note that the attacker also knows ${\partial \loss_1}/{\partial b^{(1)}_i}$ for $b^{(1)}=\hat b^{(1)}_{i,1} $, i.e., when $\x_1$ activates neuron~$i$.
%$\frac{\partial \loss_1}{\partial b^{(1)}_i}\rvert_{b^{(1)}_i = \hat b^{(1)}_{i,1}}$. 
% The attacker can For subsequent inputs, the attacker must first compute the $\alphas$ coefficients, and then reconstruct the corresponding data points. 
Suppose now that the attacker already reconstructed the first $k$ inputs $\x_1, \dots, \x_k$ and the corresponding partial derivatives ${\partial \loss_1}/{\partial b^{(1)}_i}, \dots, {\partial \loss_k}/{\partial b^{(1)}_i}$ when the corresponding inputs activate neuron $i$.
Let $h_k$ denote the partial derivative of the loss for $\hat{b}^{(1)}_{i,k}$, i.e., $h_k ={\partial \loss }/{\partial b^{(1)}_{i}}$.
Due to \eqref{eq:aggr_coeff} and Ass.~\ref{ass:grad}, the attacker can reconstruct ${\partial \loss_{k+1}}/{\partial b^{(1)}_i}$ by
\begin{equation}
\frac{\partial \loss_{k+1}}{\partial b^{(1)}_i}= n (h_{k+1} - h_k).
\end{equation}
The attacker can then compute the set of coefficients $\alpha_{j,k+1}$ in the linear combination of $\g_{k+1}$ in \eqref{eq:alpha_derivation} and reconstruct $\x_{k+1}$ by
%Suppose he has already reconstructed the first $k$ inputs $x_1, \dots, x_k$, which are separated by the first $k+1$ hyperplanes, and that he has determined their associated gradients, $\frac{\partial\loss_1}{\partial\zini}, \dots, \frac{\partial\loss_k}{\partial\zini}$. The objective is to reconstruct $x_{k+1}$, that is the input located between the $(k+1)$-th and $(k+2)$-th hyperplanes. According to (\ref{eq:obs_derivation}), the observed aggregated gradient at this stage is given by:  
% \begin{equation*} 
% \g_{k+1} =\sum_{j=1}^{k+1} \alpha_j \x_j.
% \end{equation*}
% To compute the $\alpha$ coefficients, the attacker first recovers $\frac{\partial{\loss_{k+1}}}{\partial{z_{i}^{(1)}}}$ as follows:
% \begin{equation*}
%     \frac{\partial{\loss_{k+1}}}{\partial{z_{i}^{(1)}}}=\sum_{j=1}^{k+1}\frac{\partial{\loss_j}}{\partial{z_{i}^{(1)}}} - \sum_{j=1}^{k}\frac{\partial{\loss_j}}{\partial{z_{i}^{(1)}}}.
% \end{equation*}
% This operation is only valid under Assumption~\ref{ass:grad}, which ensures that the gradients $\frac{\partial{\loss_j}}{\partial{z_{i}^{(1)}}}$ remain unchanged for all the previous $k$ inputs in the two observations. Subsequently, by applying \eqref{eq:alpha_derivation}, it is possible to derive the values of $\alpha_1, \dots, \alpha_{k+1}$. Finally, the attacker reconstructs $x_{k+1}$ by
\begin{equation}\label{eq:input_rec}
\x_{k+1} = \frac{\g_{{k}+1} - \sum_{j=1}^{k} \alpha_{j,k+1} \x_j}{\alpha_{j,k+1}}.
\end{equation}
%relying on the previously reconstructed points $x_1, \dots, x_k$, and the newly computed coefficients $\alpha_1, \dots, \alpha_{k+1}$.
%By repeating this process inductively, the attacker can recover the entire batch.

While we have described the process in two consecutive phases---first isolating the inputs with the hyperplanes and then reconstructing the inputs---we note that these procedures can be intertwined. In particular, inputs $\x_1, \dots, \x_k$ can be reconstructed as soon as the corresponding bias values $\hat b^{(1)}_{i,1}, \dots, \hat b^{(1)}_{i,k}$ have been identified.

\subsection{Controlling the Gradients}\label{sec:control_weights}

Assumption~\ref{ass:grad} requires  that the gradients ${\partial{\loss_j}}/{\partial{b^{(1)}_i}}$ are constant across attack rounds (i.e., for different values of~$b^{(1)}_{i}$, as far as the corresponding input activates neuron~$i$).
Here we show how we can satisfy this assumption %the assumption in an FC-NN 
for any classification task over the set of classes $\mathcal{C}$.
Consider a model that employs a standard cross-entropy loss function. Given a sample $(\x_j, y_j)$, we have
\begin{equation*}
    \loss_j=-\log \left({\frac{\exp{{(z^{(2)}_{y_j})}}}{\sum_{c\in \C}\exp{(z^{(2)}_{c})}}}\right).
\end{equation*}
Simple calculations (App.~\ref{app:partial_derivative}) show that if image $j$ activates neuron $i$, then
\begin{flalign}
\label{eq:part_loss_bias}
    &\frac{\partial{\loss_j}}{\partial{b^{(1)}_i}} = - w_{y_j, i}^{(2)} + \sum_{\substack{k \in \C }}w_{k,i}^{(2)}  \frac{\exp{(z^{(2)}_k)}}{\sum_{c\in \C}\exp{(z^{(2)}_c)}},
    \end{flalign}
where $z^{(2)}_k = \W^{(2)}_k(\relu(\W^{(1)}\x_j + \b^{(1)}))+ b^{(2)}_k$.
If we now set large enough bias values in the second layer, we obtain $\frac{\exp{(z^{(2)}_k)}}{\sum_{c\in \C}\exp{(z^{(2)}_c)}} \approx \frac{1}{|\C|}$, and then 
%We can now explain how to control the gradients $\frac{\partial{\loss_j}}{\partial{z^{(1)}_i}}$.
% \andre{I removed a whole part here that was already covered in 4.2 in my opinion.}
% Recall that our attack relies on finding two consecutive hyperplanes that enclose each input. This can be accomplished by assigning different $\bin$ values at each communication round. However, varying the input bias values affects the gradients $\frac{\partial{\loss_j}}{\partial{\b^{(1)}}}$, causing them to fluctuate across rounds. As a result, the hyperplanes' positions cannot be reliably used to precisely locate the inputs, as the condition $\g_{k-1} = \g_{k}$ would not be sufficient to detect a point.
%Let us fix all model parameters except for the input bias terms $\bin$, which are updated at each round to perform the binary search. As shown in \eqref{eq:part_loss_bias}, modifying $\bin$ results in different gradient values $\frac{\partial{\loss_j}}{\partial{z^{(1)}_i}}$ for the same input $\x_j$ at each communication round. This variation prevents the attacker from reliably identifying the presence of points between hyperplanes solely by analyzing differences in the corresponding gradients, as discussed in Section~\ref{sec:loc_and_rec}.
%However, the softmax term in \eqref{eq:part_loss_bias} can be controlled by assigning a sufficiently large classification bias value, ensuring that
\begin{equation}\label{eq:control_grad}
    \frac{\partial{\loss_j}}{\partial{b^{(1)}_i}}\approx -w_{y_j, i}^{(2)} + \sum_{\substack{k \in \C }}w_{k,i}^{(2)} \frac{1}{|\mathcal{C}|}.
\end{equation}
%In this way, the partial derivatives are made independent of $b_i^{(1)}$ as required by Assumption~\ref{ass:grad}. While this reasoning holds for any choice of the matrix $\W^{(2)}$, we need ${\partial{\loss_j}}/{\partial{b^{(1)}_i}} \neq 0$, for input $\x_j$ to have a non-zero coefficient in the aggregate gradient. Moreover, for the efficient parallel search we are going to describe in the next section, we want ${\partial{\loss_j}}/{\partial{b^{(1)}_{i_1}}} = {\partial{\loss_j}}/{\partial{b^{(1)}_{i_2}}}$ for all neurons, and then the rows of $\W^{(2)}$ to be identical.

In this way, the partial derivatives become independent of~$b^{(1)}_i$ and $\alpha_j$ can be artificially set to a non-negligible value without knowledge of the $y_j$.% as required by Assumption~\ref{ass:grad}. 
While this reasoning holds for any choice of the matrix $\W^{(2)}$, we require ${\partial \loss_j}/{\partial b^{(1)}_i} \neq 0$ for the input $\x_j$ to have a non-zero coefficient in the aggregate gradient. Furthermore, for the efficient parallel search described in the next section, we want ${\partial \loss_j}/{\partial b^{(1)}_{i_1}} = {\partial \loss_j}/{\partial b^{(1)}_{i_2}}$ across all neurons, which requires the columns of $\W^{(2)}$ to be identical. These constraints are easily satisfied since the attacker has control over all model parameters.

\subsection{Parallel Search}
\label{sec:parallel}
% old version
% Here we present the final algorithm. In the preceding sections, we considered the case of a single neuron in the first fully connected (FC) layer. However, the attack can be parallelized to utilize all $N$ neurons in the layer, significantly reducing the number of communication rounds. This is achieved by assigning the same random direction to each row of $\Win$ and incremental values to $\bin$.
% The overall strategy involves initially placing equally spaced hyperplanes and then refining the search intervals based on the location of inputs. By fixing the contribution of each input to the gradient computation in \eqref{eq:aggr_obs}, it becomes straightforward to identify intervals that do not contain any input, as discussed in Section \ref{sec:}. Specifically, if two consecutive hyperplanes do not isolate any input, this can be determined by checking whether $\g_k$ and $\g_{k+1}$ are identical. When this occurs, the space between the two hyperplanes can be safely excluded from the search space, thereby narrowing the intervals more efficiently. Algorithm~\ref{algo:parallelattack} implements the full solution.

% \francesco{There is a missing step in Algo 1: the actual reconstruction part of \eqref{eq:input_rec}. I think we should add it.}

We now present the complete version of our attack. So far, we have described a method 
in which the attacker tests a different bias value for neuron $i$ in different rounds. 
%in the case of a single neuron.
This approach can be parallelized across all the $N$ neurons in the layer, significantly reducing the number of required communication rounds.
The procedure is described in Algorithm~\ref{algo:parallelattack}.

The initial search space for the bias values, $[l_1, u_1]$, is determined based on the range of possible input feature values. In the first communication round, this interval is divided into $N$ equal-length subintervals. The procedure UpdateSearchState in Algorithm~\ref{algo:update_search} identifies which subintervals require further exploration. Subintervals containing no inputs, as well as those whose size becomes smaller than the threshold~$\varepsilon$, are discarded.

At each round, $\mathcal{I}$ represents the set of remaining subintervals. The attacker tests $N$ new bias values within these subintervals, distributing the values roughly evenly among them. More precisely, if $N$ cannot be evenly divided among the subintervals, the extra bias values are assigned to the longer intervals. This procedure, called UpdateHyperplanes, is detailed in Algorithm.~\ref{algo:update_hp}.

\begin{algorithm}[t]
\caption{Parallel attack}\label{algo:parallelattack}
\textbf{Input}: 
 $[l_1, u_1]$ initial search space of bias values, the set of attack rounds $\mathcal{T}=\{1,...,T\}$
\begin{algorithmic}[1]
    \STATE $\obs \leftarrow \{\}$
    \STATE $\mathcal I \leftarrow \{[l_1, u_1]\}$ 
    %\STATE Draw a hyperplane $w\in \real^d$ and a vector $v\in \real^C$, set each row of $\Win$ to $w$ and each column of $\Wout$ to $v$
    \STATE Draw a vector $\mathbf{w}\in \real^d$ and a vector $\mathbf{v}\in \real^C$, set each row of $\Win$ to $\mathbf{w}$ and each column of $\Wout$ to $\mathbf{v}$
    \STATE Set all entries of $\bout$ to the same large value
    %$\bout$ set to large equal values
    \FOR{$t \in \{1,...,T\} $ }
        \STATE $\bin$ $\leftarrow \mathrm{UpdateHyperplanes}(\mathcal{I})$\label{line:udpate_hyperplane}
        \STATE server sends malicious parameters $\theta$ to the client
        \STATE server receives gradient updates $\frac{\partial{\loss}}{\partial{\W^{(1)}}}$ and $\frac{\partial{\loss}}{\partial{\bin}}$
        \STATE server computes $\g_i$ using \eqref{eq:obs_derivation}, $\forall i \in \{1,...,N\}$
        \label{line:computeg}
        \STATE $\obs \leftarrow \obs \cup \{(\gi, \frac{\partial{\loss}}{\partial{b^{(1)}_i}}, b_i^{(1)}), \forall i = \{1, ..., N\}\}$ \label{line:updatestrip}
        \STATE $\obs, \mathcal{I}\leftarrow \mathrm{UpdateSearchState}(\mathcal{G})$
    \ENDFOR
    \STATE Reconstruct the inputs from $\obs$ using \eqref{eq:input_rec} \label{line:reconstruction}
\end{algorithmic}
\end{algorithm}

\begin{algorithm}[h]
\caption{UpdateSearchState}\label{algo:update_search}
\textbf{Input}: The set of strips $\obs$.

\begin{algorithmic}[1]
\STATE $\intervals \leftarrow \{\},\; \obsnew \leftarrow \{\}$
\STATE sort $\obs$ by ascending value of the bias values 
\FOR {$i = 2, \dots, |\mathcal{G}|$}
    \IF {$\gi \neq \mathbf{g}_{i-1}$ \AND $b_i - b_{i-1} \ge \varepsilon$ }
        \STATE $\obsnew \leftarrow \obsnew \cup \{\left(\gi, \frac{\partial{\loss}}{\partial{b^{(1)}_i}}, \bini\right) \}$
        \STATE $\intervals \leftarrow \intervals \cup \{[b_{i-1}^{(1)}, \bini]\}$
    \ENDIF
\ENDFOR
\STATE Return $\obsnew, \intervals$
\end{algorithmic}
\end{algorithm}

\begin{algorithm}[h]
\caption{UpdateHyperplanes}\label{algo:update_hp}
\textbf{Input}: set of search intervals $\mathcal{I} = \{[l_1, u_1],[l_2, u_2],\ldots, [l_k, u_k],\ldots\}$

\begin{algorithmic}[1]
    \STATE $M = |\mathcal I|$
    \IF {$M =  1$}
        \STATE $\bin \leftarrow \big\{ l_1 + i \frac{u_1 - l_1}{N+1} | i = 1, \dots, N \big\}$
    \ELSE
        \STATE sort $\intervals$ by descending length $(u_k - l_k)$
        \STATE $r \leftarrow N \bmod M,\; q \leftarrow \lfloor N/M\rfloor,\; j \leftarrow 0$
        \FOR {$k=1,...,M$}
            \IF{$k < r$}

                \STATE $\bin_{j:j+q} \leftarrow\{ l_k + i \frac{u_k - l_k}{q + 2} | i = 1, \dots, q+1 \}$
                \STATE $j \leftarrow j + q + 1$
                \ELSE
                \STATE $\bin_{j:j+q-1} \leftarrow\{ l_k + i \frac{u_k - l_k}{q+1} | i = 1, \dots, q \}$
                \STATE $j \leftarrow j + q$
            \ENDIF            
        \ENDFOR
    \ENDIF
    \STATE Return $\bin$
\end{algorithmic}
\end{algorithm}

%STATE $\obs(t_i) = \obs(t_{i-1}) \cup~((\gi, \gb, b_i),...,(\mathbf{G}_{\mathbf{W}_{N}}, G_{b_N}, b_N))$.

%%%% OLD
% \andre{Just a small sketch to remember for now.}
% We assume $N+1 \geq 2n$.
% Note that some assumption of this type is necessary as if there are too many images to reconstruct compared to $N$, then we inevitably need linear time in $n$.
% We have a total width of $W$ to search, and we are done when the total width of the active strips reach $n\eps$.
% At each step we have at most $n$ active strips, which we recurse on, out of $N+1$ that we probed.
% Hence we have a reduction of width that we search in by a factor of $\frac{N+1}{n}$ each round.
% Combining the above statements, we end up with a number of rounds of
% \[
%     \left\lceil \log_{\frac{N+1}{n}} \frac{W}{n\eps} \right\rceil \leq \log_2 \left(\frac{W}{n\eps}\right) + 1.
% \]

\subsection{Complexity Analysis}

To evaluate the time complexity of the attack (Alg.~\ref{algo:parallelattack}), we decompose the operations performed by the server during each attack round, and subsequently analyze the total computational cost. 

At each round, the server first prepares the malicious hyperplanes for the targeted client (line~6). This procedure requires $\mathcal{O}(n\log n)$ operations, primarily due to the sorting operation on the set of subintervals $\intervals$ of size at most $n$ (line~5 in Alg.~\ref{algo:update_hp}).\footnote{As we can see from Alg.~\ref{algo:update_search}, $\intervals$ and $\mathcal{G}_{new}$ consist of non-overlapping intervals, each containing at least one point. Therefore, given a batch size of $n$, we have $|\intervals| \leq n$ and $|\mathcal{G}_{new}| \leq n$.}
After receiving the gradients from clients, the server then computes $N$ observations $\g_i$ of size $d$, which requires $\mathcal{O}(Nd)$ operations, and adds them to the current set of strips $\mathcal{G}$ (lines~9 and~10). After this step, the size of $\mathcal{G}$ becomes at most $n+N$. To prepare the next-round attack (Alg.~\ref{algo:update_search}), the server updates the search space over the strips, which involves sorting the set $\mathcal{G}$ and iterating over it to compare vectors $\g_i$. This step incurs a computational cost of $\mathcal{O}(\max((N+n)\log(n+N), (n+N)d))$. 

We then analyze the number of rounds $T$ required to isolate all inputs (i.e., achieve full batch recovery) with high probability. Assuming that the hyperplane $\mathbf{w}$ is drawn from a standard normal distribution and that $N \geq n$, we show that $T$ is $\mathcal{O}(\log \frac{n}{N})$ when the threshold $\epsilon$ is set to $\mathcal{O}(1/n^2)$ (Appendix.~\ref{app:bound_on_rounds}).

The final reconstruction step requires an additional $\mathcal{O}(nd)$ operations (line~12). With hardware optimized for parallel computation, the dimensional dependency on $d$ in the above steps can be effectively mitigated, resulting in a total complexity of  $\mathcal{O}\left(\log\frac{n}{N}((N+n)\log(n+N))\right)$.

\section{Experiments}

\begin{figure*}[tb]
    \centering
    % First row
    \subfloat[ImageNet\label{fig:batch_eff_imagenet}]{%
        \includegraphics[width=0.31\linewidth]{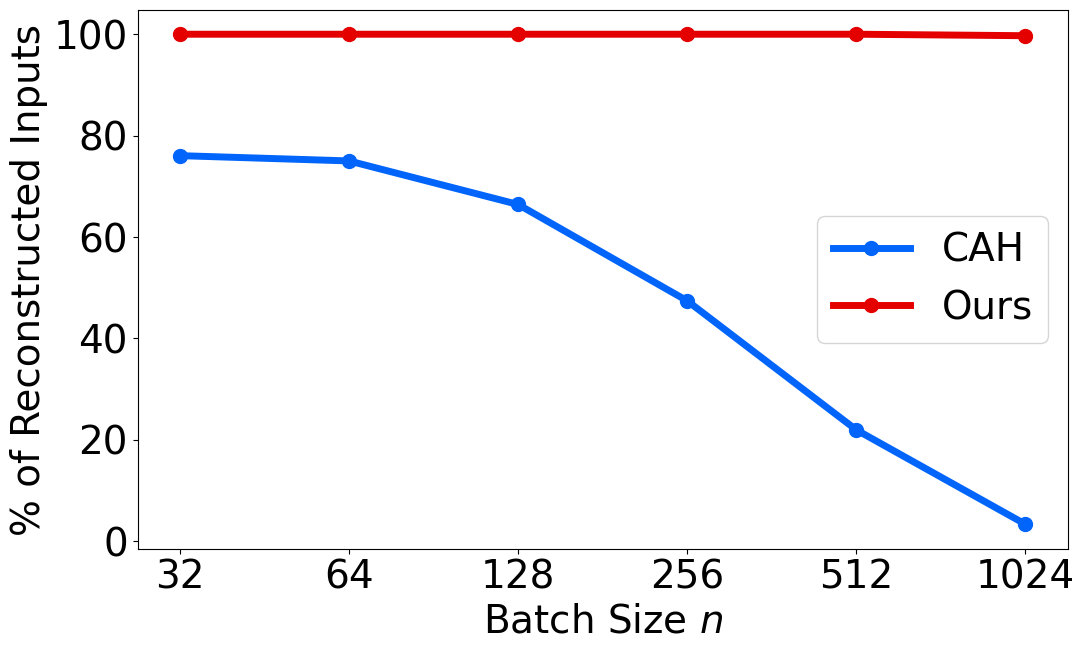}
    }\hspace{0.05\linewidth}
    \subfloat[HARUS\label{fig:batch_eff_harus}]{%
        \includegraphics[width=0.31\linewidth]{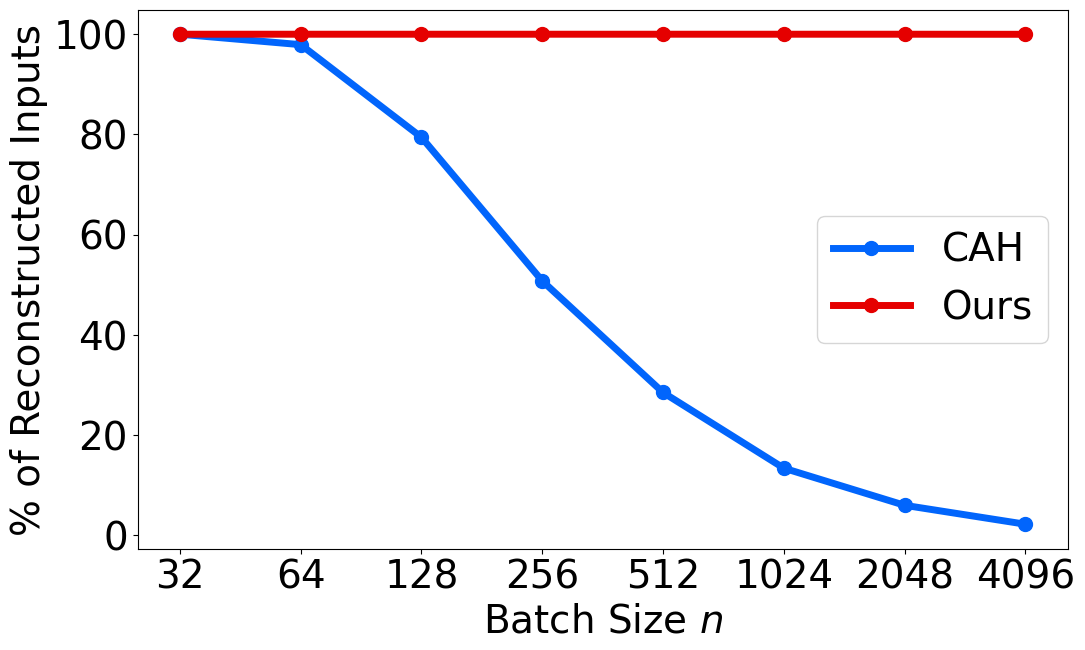}
    }

    % \vskip\baselineskip % Add vertical space between rows

    % Second row
    \subfloat[ImageNet\label{fig:rounds_eff_imagenet}]{%
        \includegraphics[width=0.31\linewidth]{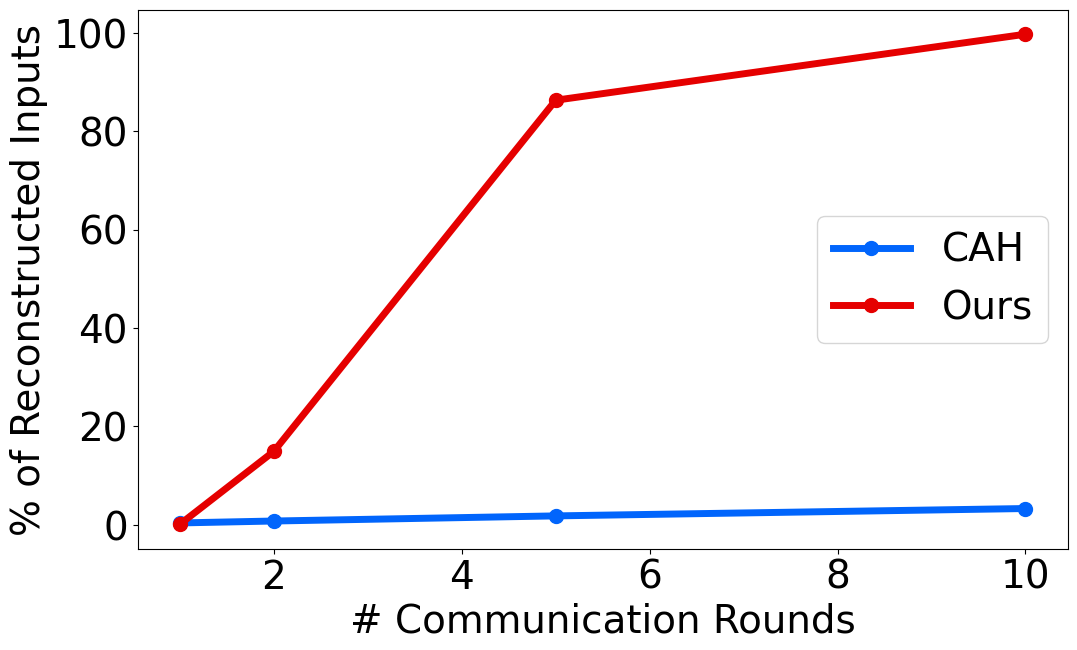}
    }\hspace{0.05\linewidth}
    \subfloat[HARUS\label{fig:rounds_eff_harus}]{%
        \includegraphics[width=0.31\linewidth]{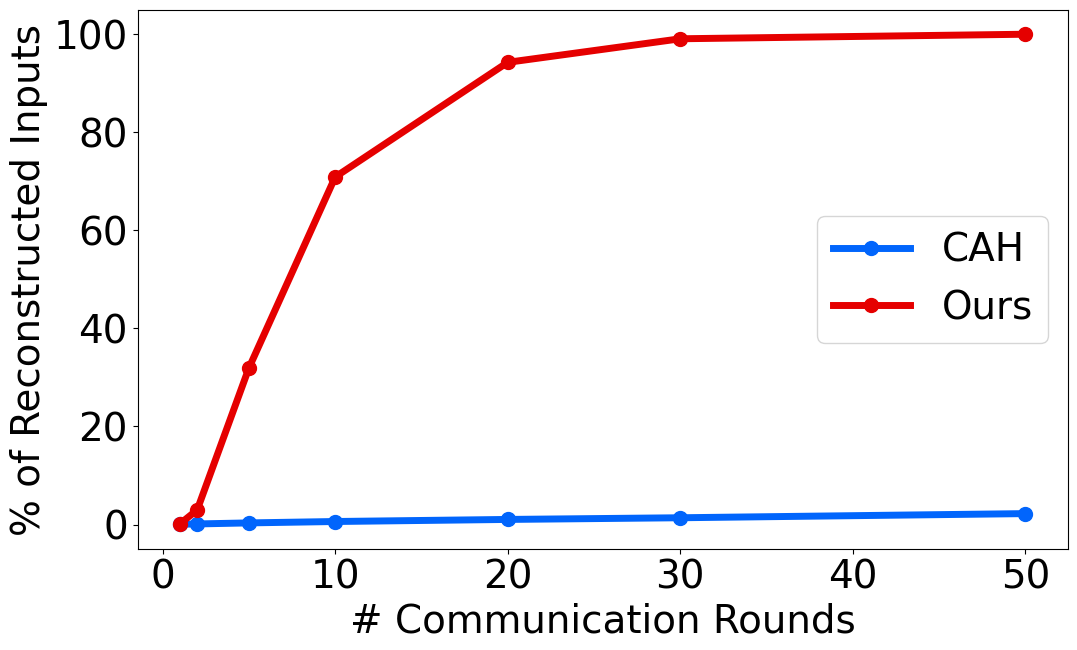}
    }

    \caption{The percentage of perfectly reconstructed inputs on ImageNet and HARUS for a two-layer FC NN with $N=1000$ neurons in the first layer. In \ref{fig:batch_eff_imagenet} the reconstruction is evaluated after 10 communication rounds, while in \ref{fig:batch_eff_harus} it is measured after 50 rounds. Figure~\ref{fig:rounds_eff_imagenet} illustrates the impact of the number of communication rounds for batch size $n=1024$, and \ref{fig:rounds_eff_harus}  for $n=4096$. }
    \label{fig:main_fig}
\end{figure*}

We conduct experiments to demonstrate the efficacy of our proposed 
%malicious weight modifications 
attack on both image and tabular data \footnote{Code: \RaggedRight\url{https://github.com/francescodiana99/cutting-through-privacy}}. Our evaluation encompasses two distinct datasets, each representing a different data modality. 
For image classification tasks, we use the ImageNet ILSVRC 2012 dataset \citep{imagenet}, that comprises a total of 1000 classes. As preprocessing steps, all images are rescaled to the $[0,1]$ range and resized to $224 \times 224$ pixels. 
To evaluate performance on tabular data, we test the Human Activity Recognition Using Smartphones dataset \citep{harus} (HARUS), which includes recordings of 30 subjects performing various activities while carrying a waist-mounted smartphone equipped with embedded sensors. The dataset comprises 561 features and 6 classification labels, representing the activity performed by each user. All features are scaled between $-1$ and $1$.
Each client trains an %fully-connected neural network
FC-NN with two layers. Unless stated otherwise, the first layer consists of $N=1000$ neurons.
For all the experiments, we assume that training is performed using FedSGD with full-batch updates. An image is considered reconstructed if it has SSIM $\geq 0.99$ %or L2-norm difference $<0.1$ 
relative to the true image (Figures~\ref{fig:images_examples1} and~\ref{fig:images_examples2} provide examples of reconstructed images). For tabular data, we consider an input to be perfectly reconstructed if the L2-norm difference between the true and recovered data point is $<0.1$. 

We compare our method to the \emph{Curious Abandon Honesty} (CAH) attack \citep{curious}, allowing, for a fair evaluation, their attack to be performed across multiple communication rounds.  Each experiment is repeated three times with different random seeds to ensure consistency and reliability of the results. Details on the configuration of both attacks 
%model's parameter initialization for both the attacks 
can be found in App.~\ref{app:model_init}. 
% We compare our method to the \emph{Curious Abandon Honesty} (CAH) attack \citep{curious}, supposing, to have a fair evaluation, that their attack can be per for multiple communication rounds.
%
% Each experiment is repeated three times with different random seeds to ensure consistency and reliability of the results. 

\subsection{Experimental Results}\label{sec:exp_results}

\begin{figure}[tb]
    \centering
    \includegraphics[width=1\linewidth]{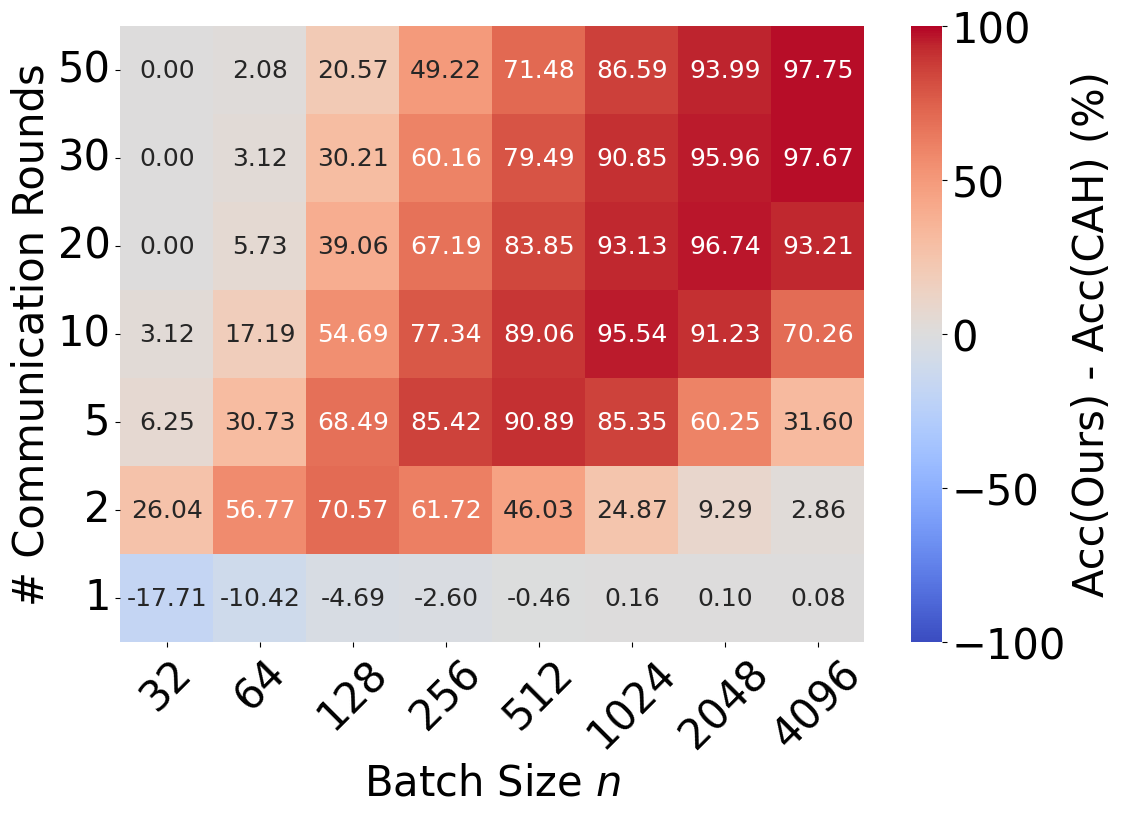}
    % \caption{Attack accuracy difference on HARUS dataset. \francesco{Think about a better description}}
    \caption{Difference in the percentage of correctly reconstructed inputs rate between our attack and the CAH attack on the HARUS dataset.
    %Accuracy difference Difference between percentage of reconstructed inputs on HARUS (i.e., a value of 100 means that we reconstruct the whole input, while the baseline reconstructs nothing). 
    %Red indicates cases where our attack achieves better reconstruction, while blue highlights instances where the baseline attack performs better.
    }
    \label{fig:heatmap_harus}
\end{figure}

The experimental results demonstrate the superior performance of our attack over the baseline across both ImageNet and HARUS datasets. On ImageNet, after 10 communication rounds, our attack consistently outperforms the baseline across all tested batch sizes, successfully recovering the entire batch in every scenario, see Figure~\ref{fig:batch_eff_imagenet}. Similarly, on HARUS, our attack achieves significant improvements, recovering 4,096 data points after 50 rounds, see Figure~\ref{fig:batch_eff_harus}.  %while the baseline method recovers less than 3\% of the dataset, see Figure~\ref{fig:batch_eff_harus}. 
%A particularly noteworthy finding is CAH limitation in handling large batches. 
On both datasets, we observe a decrease in the fraction of  samples reconstructed by CAH as the batch size increases. This demonstrates the baseline's limitations in recovering large batches when the input dimension is held constant, which is in accordance with our theoretical findings in Sec.~\ref{sec:limit}.

% We observe in Figures~\ref{fig:rounds_eff_imagenet} and \ref{fig:rounds_eff_harus} the benefits of the parallel search. The initial percentage of recovered inputs does not exceed 2\% of the full batch, but increasing the number of communication round substantially improves the recovery rate. On the contrary, the baseline does not obtain great benefits from the additional number of rounds, as their search method is randomized, and does not rely on a specific strategy to target each sample.

Figures~\ref{fig:rounds_eff_imagenet} and \ref{fig:rounds_eff_harus} further highlight the advantages of our attack and its parallel search strategy. Initially, the percentage of recovered inputs remains below 2\% of the full batch for both the attacks but, as the number of communication rounds increases, the success rate of our method improves significantly. In contrast, the baseline shows only marginal gains, as it relies on a randomized search approach.

Figure \ref{fig:heatmap_harus} illustrates the performance difference between our attack and CAH on the HARUS dataset. %The baseline exhibits better performance after a single communication round for small batch sizes due to its trap weights initialization, but at the cost of requiring additional data to tune the hyperparameter $s$. 
Two key trends emerge from our results. 
%First, the difference in performance between our attack and the baseline 
First, the performance gap increases with both the number of communication rounds and the batch size. This is clearly visible in the figure, with the largest performance difference ($97.75$ percentage points) observed at 50 rounds and a batch size of 4096. Second, different  regions of the heatmap reveal the relative strengths of each approach. % under different conditions. 
%The top-left (small batches, many rounds) suggests the baseline's effectiveness on smaller batch, and the bottom row indicates CAH's advantage in the one-shot attack, which can be attributed to their trap weights initialization, that requires additional data for tuning its hyperparameter~$s$ (see App.~\ref{app:model_init}). 
The top-left region (small batches, many rounds) suggests CAH's effectiveness on smaller batches; the bottom row highlights its advantage in one-shot attacks. However, this advantage relies on their trap weights initialization, which requires the attacker to have access to some representative data to tune some hyperparameters (see App.~\ref{app:model_init}). This relative advantage disappears as the batch size increases. The top-right region (large batches, many rounds) highlights our attack’s ability to scale, demonstrating its potential for larger batches. This supports our claim that a sufficient number of rounds enables recovery for arbitrary large batches.

We also evaluate the effect of the number of neurons $N$ in the fully connected attack layer and its relationship with attack accuracy, see Table~\ref{tab:neuron_effect}.
% From Table~\ref{tab:neuron_effect}, we observe that the percentage of reconstructed data increases with a larger number of neurons in both attacks.
%While we observe that the percentage of reconstructed data increases with a larger number of neurons in both attacks, we note that we outperform the baseline for all combinations of number of rounds and number of neurons except one; in many cases our accuracy is even about a hundred times higher.
While we observe that the percentage of reconstructed data increases with a larger number of neurons for both attacks, our attack outperforms the baseline in all but one combination of rounds and neuron counts. In many cases, our attack reconstructs nearly 100 times more inputs than CAH.
%our accuracy is nearly a hundred times higher.
We highlight a key distinction: the baseline reconstructs around 3.6\% of the input
after 50 rounds with 2000 neurons, i.e., using 100 000 different random hyperplanes.
On the other hand, our attack manages to reconstruct 38\% of the batch after only 10 rounds with only 500 neurons, i.e., using only 5000 different hyperplanes.
In comparison, our attack reconstructs more than 10 times as many input points using 20 times fewer hyperplanes than the baseline,  thanks to its efficient search procedure for correctly positioning the hyperplanes that isolate the inputs.

In terms of computational overhead, we evaluate our attack on the most computationally demanding setting—ImageNet with $n=1024$. In this case, the attack requires less than two seconds per round (primarily for lines 9 and 11 in Alg.~\ref{algo:parallelattack}) and under four seconds for the final image reconstruction phase (line 12), using an NVIDIA GeForce RTX 2080 Ti GPU. Since attacks are carried out across consecutive rounds, this additional overhead of approximately two seconds per round is negligible compared to the tens of seconds typically required to transmit the model in a single communication round.
% However, the results highlight a key distinction: after 50 rounds with 2000 neurons, the CAH attack, which explores 100,000 different random hyperplanes, achieves an accuracy three times lower than our attack, which, after just 5 rounds with 500 neurons, uses only 2,500 hyperplanes that effectively cut through the input space.
% Moreover, the table indicates that, with a sufficiently large number of neurons, even smaller networks can recover the full batch, suggesting that our attack can potentially recover the complete dataset, independently from the number of neurons in the attack layer. 

 We provide additional experimental results on HARUS, ImageNet, and CIFAR-10~\citep{cifar10} in App.~\ref{app:add_res}, including tests on a CNN architecture and extension to the multiple local steps setting, relaxing the assumption of full-batch updates. We also evaluate the effect of data heterogeneity and the effect of noise. These results confirm the same qualitative trends observed in our main results, further validating our conclusions. 
\begin{table*}[t]
    \caption{Effect of number of neurons $N$ in the reconstruction layer. Each value indicates the percentage of data exactly recovered in attacks on HARUS dataset, when $n=4096$.}
    \label{tab:neuron_effect}
    \medskip
    \centering
    \footnotesize
    \begin{tabular}{|c|c|c|c|c|c|c|c|c|}
    \hline
    \textbf{\# Rounds} & \multicolumn{7}{|c|}{\textbf{\# Neurons}}\\
    \hline
          &  & 100& 200 & 500 & 1000 & 1500 & 2000 \\
         \hline
         \multirow{2}{*}{1}& Ours &0.00\scriptsize{$\pm$0.00} &0.2\scriptsize{$\pm$0.03} &0.05\scriptsize{$\pm$0.02} &0.15\scriptsize{$\pm$0.06} &0.15\scriptsize{$\pm$0.10}&0.21\scriptsize{$\pm$0.09}   \\ 
    & CAH &0.01\scriptsize{$\pm$0.03}  &0.02\scriptsize{$\pm$0.01} &0.04\scriptsize{$\pm$0.01} &0.07\scriptsize{$\pm$ 0.04} &0.13\scriptsize{$\pm$0.02} &0.15\scriptsize{$\pm$0.05} \\ 
    \hline
         \multirow{2}{*}{5} & Ours &1.11\scriptsize{$\pm$0.31} &2.82\scriptsize{$\pm$0.37} &9.33\scriptsize{$\pm$4.66} &31.93\scriptsize{$\pm$1.80} &49.60\scriptsize{$\pm$0.80} &64.80\scriptsize{$\pm$2.08} \\ 
    & CAH &0.07\scriptsize{$\pm$0.03} &0.08\scriptsize{$\pm$0.01} &0.20\scriptsize{$\pm$0.04} &0.33\scriptsize{$\pm$0.06} &0.50\scriptsize{$\pm$0.03} &0.66\scriptsize{$\pm$0.09} \\ 

    \hline
    \multirow{2}{*}{10} & Ours &3.11\scriptsize{$\pm$0.88} &8.33\scriptsize{$\pm$2.70} &38.02\scriptsize{$\pm$1.54} &70.87\scriptsize{$\pm$1.44} &86.54\scriptsize{$\pm$0.44} &93.17\scriptsize{$\pm$0.19} \\ 
    & CAH &0.08\scriptsize{$\pm$0.06} &0.15\scriptsize{$\pm$0.03} &0.31\scriptsize{$\pm$0.07} &0.62\scriptsize{$\pm$0.01} &0.81\scriptsize{$\pm$0.09} &1.06\scriptsize{$\pm$0.08} \\ 
    \hline
    
    \multirow{2}{*}{30} & Ours &21.10\scriptsize{$\pm$1.44} &51.64\scriptsize{$\pm$1.57} &88.38\scriptsize{$\pm$0.58} &99.06\scriptsize{$\pm$0.29} &99.85\scriptsize{$\pm$0.10} &99.98\scriptsize{$\pm$0.03} \\ 
    & CAH &0.25\scriptsize{$\pm$0.07} &0.37\scriptsize{$\pm$0.09} &0.78\scriptsize{$\pm$0.09} &1.38\scriptsize{$\pm$0.08} &1.85\scriptsize{$\pm$0.13} &2.46\scriptsize{$\pm$0.20} \\ 
    
    \hline
    \multirow{2}{*}{50} & Ours &42.52\scriptsize{$\pm$2.07} &75.36\scriptsize{$\pm$1.68} &97.84\scriptsize{$\pm$0.21} &99.98\scriptsize{$\pm$0.03} &100.00\scriptsize{$\pm$0.00} &100.00\scriptsize{$\pm$0.00} \\ 
    & CAH &0.37\scriptsize{$\pm$0.07} &0.53\scriptsize{$\pm$0.12} &1.24\scriptsize{$\pm$0.24} &2.23\scriptsize{$\pm$0.10} &2.70\scriptsize{$\pm$0.12} &3.62\scriptsize{$\pm$0.47} \\ 
    \hline
    \end{tabular}
\end{table*}

\section{Discussion}\label{sec:discussion}
% \subsection{Applicability to CNNs}
% Following the approach described in (App.~B, \cite{curious}), our work can be extended to CNN with at least two FC layers by adjusting the weights of the convolutional layer to behave as an identity function. However, pooling layers and dropout may help mitigate the attack by reducing the accuracy of our search method. Alternatively, the server could first reconstruct the input to the first FC layer, and then employ an optimization-based attack to recover the input batch, similar to the approach used in \cite{ZHANG2023119421}.

% \subsection{Defenses and Mitigation Strategies}
Secure aggregation \citep{sec_agg} is often proposed as a defense mechanism against privacy attacks, as it aggregates updates from multiple clients without revealing individual client updates to the server. However, the server still obtains the average of the gradients computed over the entire union of client datasets. With a sufficient number of communication rounds, our attack could still recover all underlying data, though not necessarily associate it with a specific client.

% Secure aggregation \citep{sec_agg} is often proposed as a defense mechanism against  privacy attacks, as it aggregates updates from multiple clients using an intermediary before forwarding them to the central server. However, in our setting, the aggregation produces a linear combination containing a larger number of inputs. Given a sufficient number of attack rounds, our attack could still recover all the underlying data (but not necessarily attribute it to a client). 

Therefore, the only robust defense against our attack that we are aware of would involve implementing local differential privacy (LDP) \citep{dp_dwork, local_dp} on the client side. By adding noise to updates, % LDP would obscure individual inputs' contributions to the aggregated gradient. This would make it significantly more challenging to detect and eventually reconstruct the inputs.
LDP alters the coefficients associated to each input in \eqref{eq:aggr_obs}, thus hindering precise localization and subsequent reconstruction of the data points, at the cost of decreased model utility. We conducted a preliminary evaluation of our attack's performance when clients add Gaussian noise to their updates. Results in App.~\ref{app:add_res} show that noise perturbation decreases the proportion of inputs that our attack can perfectly recover. Nevertheless, our method remains effective: even when clients add noise, data reconstruction is still feasible and we consistently outperform the baseline in terms of reconstruction accuracy.

\section{Future Works and Conclusion}\label{sec:conclusions}
In this work, we proposed a novel data reconstruction attack in Federated Learning for classification tasks, demonstrating that a malicious server can manipulate model parameters to perfectly recover clients' input data, regardless of the batch size. 
%Our approach assumes the use of the FedSGD protocol, where clients perform a single gradient update before communicating with the server.

%Extending this attack to settings where clients perform multiple local steps presents an interesting challenge. One potential approach could involve leveraging the server's control over client training hyperparameters by setting an exceptionally low learning rate. This would ensure that local updates to the input layer would not significantly alter the orientation of the hyperplanes, thereby preserving the feasibility of the reconstruction. A similar result can be obtained by modifying the activation function so that updates are clipped. However, handling potential updates across multiple mini-batches remains an open question.

% Exploring the applicability of the attack to other machine learning tasks and architectures represents another promising direction for future work.
Exploring the applicability of our attack to other machine learning tasks and architectures remains a promising direction for future work. Although our current approach is broadly applicable—by first recovering embeddings from the first fully connected layer and then applying a model inversion attack, following the methodology of \citet{kariyappa23a_cocktail} and \citet{ZHANG2023119421}—the resulting reconstructions would still not be perfect. Therefore, developing analytical attacks that enable accurate, end-to-end reconstruction from more complex architectures and from different tasks remains an important and promising topic for future research.

Extending this attack to settings where clients perform multiple local epochs presents also an interesting challenge. In App.~\ref{app:add_res}, we carried out a preliminary evaluation by considering a scenario with one local epoch using minibatch stochastic gradient descent.
%Our approach leverages the server's control over client-side training hyperparameters, carefully selecting values that ensure updates to the input layer do not substantially alter the orientation of the hyperplanes—thereby preserving the feasibility of the reconstruction. A similar effect can also be achieved by modifying the activation function to clip updates, as suggested by \citet{fowl2022robbing}.
However, effectively handling cumulative minibatch updates over multiple local epochs remains an open question.

\begin{acknowledgements}
    This research was supported in part by the French government, through the 3IA Côte d’Azur Investments in the Future project managed by the National Research Agency (ANR) with the reference number ANR-19-P3IA-0002 and in part by the European Network of Excellence dAIEDGE under Grant Agreement Nr. 101120726. It was also funded in part by the Groupe La Poste, sponsor of the Inria Foundation, in the framework of the FedMalin Inria Challenge and by the EU HORIZON MSCA 2023 DN project FINALITY (G.A. 101168816).
    André Nusser was supported by the French government through the France 2030 investment plan managed by the National Research Agency (ANR), as part of the Initiative of Excellence of Université Côte d’Azur under reference number ANR-15-IDEX-01.
The authors are grateful to the OPAL infrastructure from Université Côte d’Azur for providing resources and support.
\end{acknowledgements}
%The applicability of the attack to other machine learning tasks and architectures (beyond those discussed in Sec.~\ref{sec:extensions}) represents another promising direction that we leave to future works. 

% Multiple update challenge, our method works well when the learning rate is small enough, point to the Robbing 

% CNN case, point to the Honest, Zhang's paepr, two methods. 

% Defenses:  DP

% Detectability: the same as the Robb one, future work...

% \clearpage

% \section*{Impact Statement}
% This work presents an attack that compromises user privacy in federated learning by reconstructing client data from model updates. While such attacks could be misused, our goal is to raise awareness of these privacy threats, helping the community to understand the limitations of federated learning and to stipulate the development and usage of effective defense strategies to preserve user privacy.
\bibliography{ref}

\newpage

\onecolumn

\title{Cutting Through Privacy:\\ A Hyperplane-Based Data Reconstruction Attack in Federated Learning\\(Supplementary Material)}
\maketitle

% \section*{Appendices}
\appendix

\section{Proof of Theorem~\ref{theo:active_bound}}
\label{sec:proof_convex_hull}
\begin{proof}
By definition of the attack method, a point can only be reconstructed using \eqref{eq:single_input_rec} if it is linearly separated by injecting malicious weights.
Thus, the number of recovered points is upper-bounded by the number of linearly separable points in the batch $\batch_{\x}$,  which corresponds to the number of vertices on % the convex hull of $\batch$.
its convex hull.
%The expected number of vertices on the convex hull of $\batch$ % 
This value is bounded by:
\begin{enumerate}
    \item $O(n^{(d-1)/(d+1)})$ if $\batch_{\x}$ is sampled uniformly at random from a unit ball \citep{Raynaud_1970}, or
    \item $O(\log^{d-1} n)$ if $\batch_{\x}$ is sampled uniformly at random from the unit hypercube \citep{DBLP:journals/jacm/BentleyKST78}, or
    \item $O(\log^{(d-1)/2} n)$ if $\batch_{\x}$ is sampled from a normal distribution with covariance matrix $\mathbf{I}_d$ \citep{Raynaud_1970}.\qedhere
\end{enumerate}
\end{proof}

\section{Partial derivative with respect to the bias}
\label{app:partial_derivative}
By application of the chain rule, the gradient $\frac{\partial{\loss_j}}{\partial{z^{(1)}_i}}$ can be computed as
\begin{equation}
\label{eq:chain_rule}
    \frac{\partial{\loss_j}}{\partial{b^{(1)}_i}} = \frac{\partial{\loss_j}}{\partial{z^{(1)}_i}} = \frac{\partial{\loss_j}}{\partial{\z^{(2)}}} \frac{\partial{\z^{(2)}}}{\partial{z_i^{(1)}}}.
\end{equation}

Considering the $k$-th element in $\z^{(2)}$, the first term in \eqref{eq:chain_rule} is
\begin{equation}
    \label{eq:part_loss}
    \frac{\partial{\loss_j}}{\partial{z^{(2)}_k}}= \begin{cases}
        -1 + \frac{\exp{(z^{(2)}_k)}}{\sum_{c\in \C}\exp{(z^{(2)}_c)}}, & \text{if $k=y_j$} \\
        \\
        \frac{\exp{(z^{(2)}_k)}}{\sum_{c\in \C}\exp{(z^{(2)}_c)}}, & \text{if $k\neq y_j$}
    \end{cases},
\end{equation}
and the second term in \eqref{eq:chain_rule} is
\begin{equation}
\label{eq:part_z_2}
    \frac{\partial{z_k^{(2)}}}{\partial{z_i^{(1)}}} = w_{k,i}^{(2)}.
\end{equation}
% \andre{The $w_{ki}$ is somewhat undefined here. Or is this so standard that there is no need to define?}
By substituting \eqref{eq:part_loss} and \eqref{eq:part_z_2} into \eqref{eq:chain_rule}, we obtain:
% \begin{flalign}
%     \label{eq:part_loss_bias}
%     &\frac{\partial{\loss_j}}{\partial{z^{(1)}_i}} = w_{y_j i}^{(2)} \left(-1 + \frac{\exp{(z^{(2)}_{y_j})}}{\sum_{c\in \C}\exp{(z^{(2)}_c)}}\right)\notag \\ 
%     &+ \sum_{\substack{k \in \C \\ k \neq y_j}}w_{ki}^{(2)}  \frac{\exp{(z^{(2)}_k)}}{\sum_{c\in \C}\exp{(z^{(2)}_c)}}= - w_{y_ji}^{(2)}\notag  \\
%     &+ \sum_{\substack{k \in \C \\ k \neq y_j}}w_{ki}^{(2)} \frac{\exp{(\W_{k}^{(2)}(\W^{(1)}\x_j + b_i^{(1)}) + b_{k}^{(2)})}}{\sum_{c\in \C}\exp{(\W_{c}^{(2)}(\W^{(1)}\x_j + b_i^{(1)}) + b_{c}^{(2)})}}. 
%     \end{flalign}

\begin{align}
    \label{eq:part_loss_bias_ext}
    \frac{\partial{\loss_j}}{\partial{b^{(1)}_i}} &= w_{y_j, i}^{(2)} \left(-1 + \frac{\exp{(z^{(2)}_{y_j})}}{\sum_{c\in \C}\exp{(z^{(2)}_c)}}\right)+ \sum_{\substack{k \in \C \\ k \neq y_j}}w_{k,i}^{(2)}  \frac{\exp{(z^{(2)}_k)}}{\sum_{c\in \C}\exp{(z^{(2)}_c)}} \\
    &= - w_{y_j,i}^{(2)}\notag  + \sum_{\substack{k \in \C}}w_{k,i}^{(2)} \frac{\exp{(\W_{k}^{(2)}(\W^{(1)}\x_j + \bin) + b_{k}^{(2)})}}{\sum_{c\in \C}\exp{(\W_{c}^{(2)}(\W^{(1)}\x_j + \bin) + b_{c}^{(2)})}}. 
    \end{align}

\section{Modifications for Neural Networks with multiple layers}\label{app:gener_nn}

Our method can be easily adapted to any fully-connected neural network with $L$ layers with ReLU activation function. In the following, to simplify the notation we will omit the ReLU activation, assuming that the input is propagated through each layer only if the corresponding output is positive.

Let us start by considering the case of a network with two hidden layers, such that 
\begin{align*}
    \z^{(1)}&=\relu(\Win \x_j+\bin), \\
    \z^{(2)}&=\relu(\W^{(2)} \z^{(1)}+ \b^{(2)}) \\
    \z^{(3)}&=\W^{(3)} \z^{(2)} + \b^{(3)},
\end{align*} where $\W^{(3)} \in \mathbb{R}^{C\times m}$, $\W^{(2)} \in \mathbb{R}^{m\times N}$, and $\W^{(1)} \in \mathbb{R}^{N\times d}$. For a fixed neuron $i$, we want to compute the gradient $\frac{\partial{\loss_j}}{\partial{b^{(1)}_i}}$ . Using the chain rule, we can express it as:
\begin{equation}
    \label{eq:chain_rule_3_layers}
    \frac{\partial{\loss_j}}{\partial{b^{(1)}_i}} = \frac{\partial{\loss_j}}{\partial{z^{(1)}_i}} = \frac{\partial{\loss_j}}{\partial{\z^{(3)}}} \cdot 
    \frac{\partial{\z^{(3)}}}{\partial{\z^{(2)}}} \cdot
    \frac{\partial{\z^{(2)}}}{\partial{z_i^{(1)}}} ,
\end{equation}

Now, substituting the result from \eqref{eq:part_loss_bias_ext} into this expression, we get:

\begin{equation}
    \label{eq:dl_db_mult}
    \frac{\partial{\loss_j}}{\partial{b^{(1)}_i}} =  {\sum_{m}}\left(\sum_{k\in \C}\left(-\mathbbm{1}_{k=y_j} + \frac{\exp{(z^{(3)}_{k})}}{\sum_{c\in \C}\exp{(z^{(3)}_c)}}\right)w_{k,m}^{(3)}\right)w_{m,i}^{(2)}, 
\end{equation}
where $\mathbbm{1}_{k=y_j}$ indicates that the term is 1 only when ${k=y_j}$, (i.e., when the class index matches the true label $y_j$).
To extend this derivation to a network with $L$ layers, we can recursively apply the chain rule and obtain:
\begin{equation}
    \frac{\partial{\loss_j}}{\partial{b^{(1)}_i}}=\sum_{m_1}...\sum_{k\in \C}\left(\gamma_k^{(L)} \prod_{l=1}^L w_{m_l, m_{l-1}} \right)w_{m_1, i},
\end{equation}
where $\gamma_k^{(L)} = \frac{\exp{(z^{(L)}_{k})}}{\sum_{c\in \C}\exp{(z^{(L)}_c)}}$, and $m_1,...,m_L$ indicate the number of rows in $\W^{(L)}$.

Therefore, by assigning identical random values to each column of the hidden layers $\W^{(2)}...\W^{(L-1)}$, we can ensure identical inputs for all classification neurons, reducing potential errors due to varying gradient contributions $\frac{\partial{\loss_j}}{\partial{b^{(1)}_i}}$.

To validate our claim, we conduct experiments on ImageNet using models with 1, 2, 3, and 4 hidden layers. The first layer has  $N=1000$, while each subsequent layer consists of 100 neurons.  We set the malicious parameters of the input and classification layers as described in App.~\ref{app:model_init}, and generate the columns of each additional hidden layer from $\mathcal{N}(0, 10^{-6})$. The corresponding bias terms are also scaled by a $10^{-3}$ factor. Table~\ref{tab:layers_effect} demonstrates that, regardless of the number of hidden layers, the reconstruction accuracy of the attack exhibits only minor fluctuations, confirming that the number of layers does not significantly affect the performance of our attack.

\begin{table}[t]
    \caption{Effect of the number of hidden layers. Each value indicates the percentage of data exactly recovered on attacks on ImageNet, when $n=512$.}
    \label{tab:layers_effect}
    \medskip
    \centering
    \begin{tabular}{|c|c |c|c|c|}
    \hline
    \textbf{\# Rounds} & \multicolumn{4}{|c|}{\textbf{\# Hidden Layers}}\\
    \hline
     &1 & 2 & 3 & 4 \\
    \hline
    1 &0.20\scriptsize{$\pm$0.00}  &0.20\scriptsize{$\pm$0.00} &0.20\scriptsize{$\pm$0.00} &0.20\scriptsize{$\pm$0.00} \\
    2 &35.61\scriptsize{$\pm$3.12} &35.16\scriptsize{$\pm$3.32} &35.29\scriptsize{$\pm$3.33} &35.29\scriptsize{$\pm$3.33} \\
    3 &78.38\scriptsize{$\pm$2.79} &77.99\scriptsize{$\pm$2.93} &78.26\scriptsize{$\pm$2.66} &78.19\scriptsize{$\pm$2.74} \\
    5 &97.14\scriptsize{$\pm$0.98}&97.01\scriptsize{$\pm$1.19} &97.14\scriptsize{$\pm$0.98} &97.14\scriptsize{$\pm$0.98} \\
    7 &99.74\scriptsize{$\pm$0.45}&99.74\scriptsize{$\pm$0.45}&99.74\scriptsize{$\pm$0.45}&99.74\scriptsize{$\pm$0.45} \\
    10 &100.00\scriptsize{$\pm$0.00} &100.00\scriptsize{$\pm$0.00} &100.00\scriptsize{$\pm$0.00} &100.00\scriptsize{$\pm$0.00} \\

    \hline
    \end{tabular}
\end{table}

\section{Bound on the Number of Rounds} \label{app:bound_on_rounds}
Assume that the number of neurons is at least the number of input points that we want to recover, i.e., $N \geq n$.
In the first step we cut the initial interval of width $W$ into intervals of width $\frac{W}{N}$.
In any later step, we cut each interval at least $\left\lfloor \frac{N}{n} \right\rfloor$ times and hence, the intervals in the next round have a width that is reduced by a factor of at least $\left\lfloor \frac{N}{n} \right\rfloor + 1$.
We stop when the interval width reaches $\eps$.
Hence, the number of steps is bounded by
\begin{equation}\label{eq:bound_rounds}
\left\lceil \log_{\lfloor \frac{N}{n} \rfloor + 1} \left( \frac{W}{N \cdot \eps} \right) \right\rceil + 1 \leq \left\lceil\log_2 \left( \frac{W}{N \cdot \eps} \right) \right\rceil + 1.
\end{equation}
%The term on the right-hand side represents the number of rounds required for a simple binary search, as would be the case if the attacker were to consider only a single neuron. 
It is easy to show that this analysis is tight by an example where each interval initially contains a single image.

% Let us finally give an intuition about how to choose the value for $\varepsilon$.
% Let $\delta$ be the minimum distance that any two input points have.
% By standard calculations, one can show that their distance projected to a random direction is at least $\delta/d$ in expectation, where $d$ is the dimensionality of the inputs.
% Assuming that inputs are normalized to $[0,1]^d$, we have that the maximum distance of two inputs is $d$.
% Hence, using an argument similar to Markov's inequality, we obtain that the distance after projection of two inputs is at least in the order of $\delta/d^3$ with constant probability.
% Finally, applying a union bound over all pairs of inputs, we can set $\varepsilon = \frac{\delta}{d^3 n^2}$.
% Hence, the number of rounds is upper bounded in the order of $\log_2\left(\frac{W d n}{N \delta} \right)$.

Finally, we show how to select the value for $\varepsilon$. Let $\batch_{\x}=\{\x_1,\dots,\x_n\}$ be the set of samples that the attacker wants to reconstruct. Assume that inputs have a minimum distance 
\begin{equation}
    \Delta=\min_{\substack{i,j \\ i \neq j}}\lVert \x_i - \x_j \rVert.
\end{equation}
Draw a random direction vector $\mathbf{v} \sim \mathcal{N}(0, \mathbf{I}_{d\times d})$. The distance of two projections along $\mathbf{v}$ of points in $\batch_{\x}$ will be the random variable
\begin{equation}
   \mathrm{p}_{\mathbf{v}}(\x_i, \x_j)= \langle \x_i-\x_j, \mathbf{v}\rangle \sim \mathcal{N}(0, \lVert \x_i - \x_j \rVert).
\end{equation}
Therefore, we can compute:
\begin{align}
    \prob\left(\lvert \mathrm{p}_{\mathbf{v}}(\x_i, \x_j) < \varepsilon \rvert\right)&= \prob\left(q\in \left[\frac{-\varepsilon}{\lVert \x_i - \x_j \rVert}, \frac{\varepsilon}{\lVert \x_i - \x_j \rVert}\right]\right) \\&\leq \prob\left(q \in \left[\frac{-\varepsilon}{\Delta}, \frac{\varepsilon}{\Delta}\right]\right) \\
    &\leq \frac{1}{\sqrt{2\pi}}2\frac{\varepsilon}{\Delta} = \sqrt{\frac{2}{\pi}}\frac{\varepsilon}{\Delta}, 
\end{align}
where $q \sim \mathcal{N}\left(0, 1\right)$. Here, the last inequality uses the fact that the Gaussian pdf is maximized at 0 and our integration interval is symmetric around 0. Then, applying the union bound, we get:

\begin{align}
    \prob\left(\exists \ i,j \ \land \ i \neq j:  \lvert \mathrm{p}_{\mathbf{v}}(\x_i, \x_j) < \varepsilon \rvert\right) &= \prob\left( \underset{\substack{i,j \\ i\neq j}}{\bigcup}\left\{\lvert \mathrm{p}_{\mathbf{v}}(\x_i, \x_j) < \varepsilon \rvert\right\} \right ) \\
    &\leq \sum_{\substack{i,j \\ i\neq j}} \prob\left(\lvert \mathrm{p}_{\mathbf{v}}(\x_i, \x_j) < \varepsilon \rvert\ \right) \\
    &\leq \frac{n^2}{2}\sqrt{\frac{2}{\pi}}\frac{\varepsilon}{\Delta}.
\end{align}

Hence, by selecting $\varepsilon=\sqrt{2\pi}\frac{\Delta}{n^2}\delta$, all the projections will be at least $\varepsilon$-distant with high probability. After substituting the value of $\varepsilon$ in \eqref{eq:bound_rounds}, the number of rounds becomes:
\begin{equation}\label{eq:final_rounds_bound}
\left\lceil \log_{\lfloor \frac{N}{n} \rfloor + 1} \left( \frac{Wn^2}{\sqrt{2\pi}N\Delta \delta} \right) \right\rceil + 1 \leq \left\lceil\log_2 \left( \frac{Wn^2}{\sqrt{2\pi}N\Delta \delta} \right) \right\rceil + 1,
\end{equation}
with probability at least $1-\delta$.

% TODO: this should be resolved before the camera-ready!
% \andre{Short argument why if two images are far, then also they are probably far when projected to a random dimension?}
% \andre{Write something about the choice of $\varepsilon$.}
% \section{Update Rule for Parallel Search}\label{app:update_rule}
% Algorithm \ref{algo:update_hp} describes how to place the hyperplane in our parallel binary search, proposed in Section~\ref{sec:parallel}. 
% \begin{algorithm}[h]
% \caption{UpdateHyperplanes}\label{algo:update_hp}
% \textbf{Input}: set of search intervals $\mathcal{I}$, model input bias $\bin$

% \begin{algorithmic}[1]
%     \STATE $M = |\mathcal I|$
%     \IF {$M =  1$}
%         \STATE $\bin \leftarrow \big\{ l_1 + k \frac{u_1 - l_1}{N+1} | k = 1, \dots, N \big\}$
%     \ELSE
%         \STATE sort $\intervals$ by descending length $(u_k - l_k)$
%         \STATE $r \leftarrow N \bmod M,\; q \leftarrow \lfloor N/M\rfloor,\; j \leftarrow 0$
%         \FOR {$k=1,...,M$}
%             \IF{$k < r$}

%                 \STATE $\bin_{j:j+q} \leftarrow\{ l_k + i \frac{u_k - l_k}{q + 2} | i = 1, \dots, q+1 \}$
%                 \STATE $j \leftarrow j + q + 1$
%                 \ELSE
%                 \STATE $\bin_{j:j+q-1} \leftarrow\{ l_k + i \frac{u_k - l_k}{q+1} | i = 1, \dots, q \}$
%                 \STATE $j \leftarrow j + q$
%             \ENDIF            
%         \ENDFOR
%     \ENDIF
%     \STATE Sort $\bin$ in ascending order.
%     \STATE Return $\bin$
% \end{algorithmic}
% \end{algorithm}

\section{Attacks Configuration}\label{app:model_init}
% We used different initialization strategies for ImageNet and HARUS. On ImageNet, each row of the input weights $\Win$ is initialized with identical values drawn from a normal distribution $\Win_i \sim\mathcal{N}(0, 10^{-18})$, while classification weights' initial values are drawn from $\W_i^{(2)} \sim \mathcal{N}(0, 10^{-4})$. The classification bias values $\bout_i$ are set to $10^{12}$. For experiments on HARUS, we draw each row of $\Win$ from $\Win_i \sim\mathcal{N}(0, 10^{-20})$, and set $\W_i^{(2)} \sim \mathcal{N}(0, 10^{-8})$. Classification bias values $\bout_i$ are set to $10^{13}$. We chose small values for $\Win_i$ and large values for $\bout_i$ to control the softmax computation in PyTorch, since softmax in \eqref{eq:part_loss_bias} is computed as $\frac{e^{x_i}}{\sum_j e^{x_j}} = \frac{e^{-m}}{e^{-m}}\frac{e^{x_i}}{\sum_j e^{x_j}}= \frac{e^{x_i-m}}{\sum_j e^{x_j-m}}$ to improve numerical stability, where $m=\max{(x_j)}$. Therefore, setting large values for $\bout$ is not sufficient. We also need small values for$\Win_i$ and $\W_i^{(2)}$ to maintain the desired behavior.
% To test the CAH attack,
% %in \cite{curious}, 
% we draw the weights $\Win \sim \mathcal{N}(0, \frac{1}{2})$ and set the scaling factor $s=0.99$ for ImageNet experiments, following the setup described in\citep{curious}. For experiments on HARUS we use $\Win \sim \mathcal{N}(0, 1)$ and set $s=0.97$ after a tuning process.

In our attacks, we used the same initialization strategies for ImageNet and HARUS. Each row of the input weights $\Win$ is initialized with identical values drawn from a normal distribution $\Win_i \sim\mathcal{N}(0, 10^{-2})$, while classification weights' column values are drawn from $\W_i^{(2)} \sim \mathcal{N}(0, 10^{-2})$. However, the choice of the distribution does not represent a limiting factor, and different distribution can be employed. The classification bias values $\bout_i$ are set to $10^{25}$. We chose a large value for the classification bias according to \eqref{eq:control_grad}. We did not select a specific value for $\varepsilon$, instead, we iteratively selected as many bias values as were compatible with the number of communication rounds $T$ available to the attacker. Consequently, a total of $NT$ bias values were selected following Alg.~\ref{algo:update_hp}.

To test the CAH attack,
%in \cite{curious}, 
we draw the weights $\Win \sim \mathcal{N}(0, \frac{1}{2})$ and set the scaling factor $s=0.99$ for ImageNet experiments, following the setup described in \citep{curious}. For experiments on HARUS, we draw the input weights from a normal distribution $\Win \sim \mathcal{N}(0, 1)$ and select $s=0.97$ after a tuning process.

To minimize reconstruction errors arising from numerical inaccuracy, the results in the main paper are obtained using double precision. However, additional single precision experiments are presented in App.~\ref{app:add_res}.

\section{Additional experimental results}\label{app:add_res}
\subsection{CIFAR-10}
\begin{figure*}[h]
    \centering
    % First row
    \subfloat[\label{fig:batch_eff_cifar}]{%
        \includegraphics[width=0.35\linewidth]{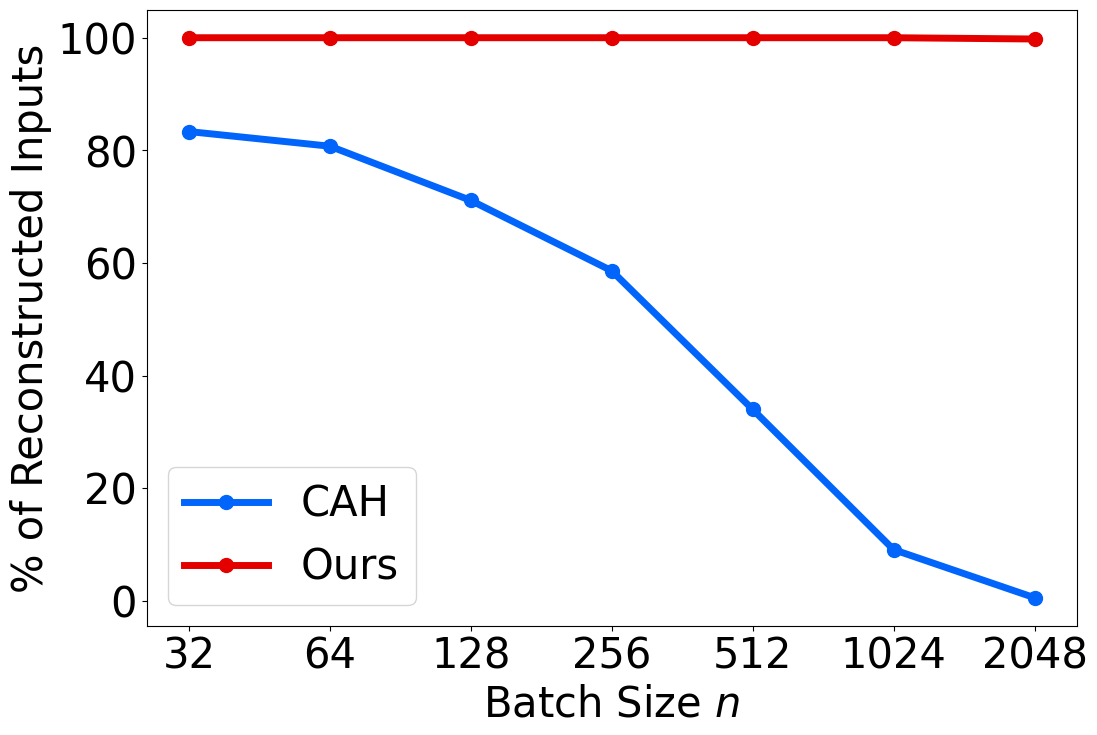}
    }\hspace{0.05\linewidth}
    \subfloat[\label{fig:rounds_eff_cifar}]{%
        \includegraphics[width=0.35\linewidth]{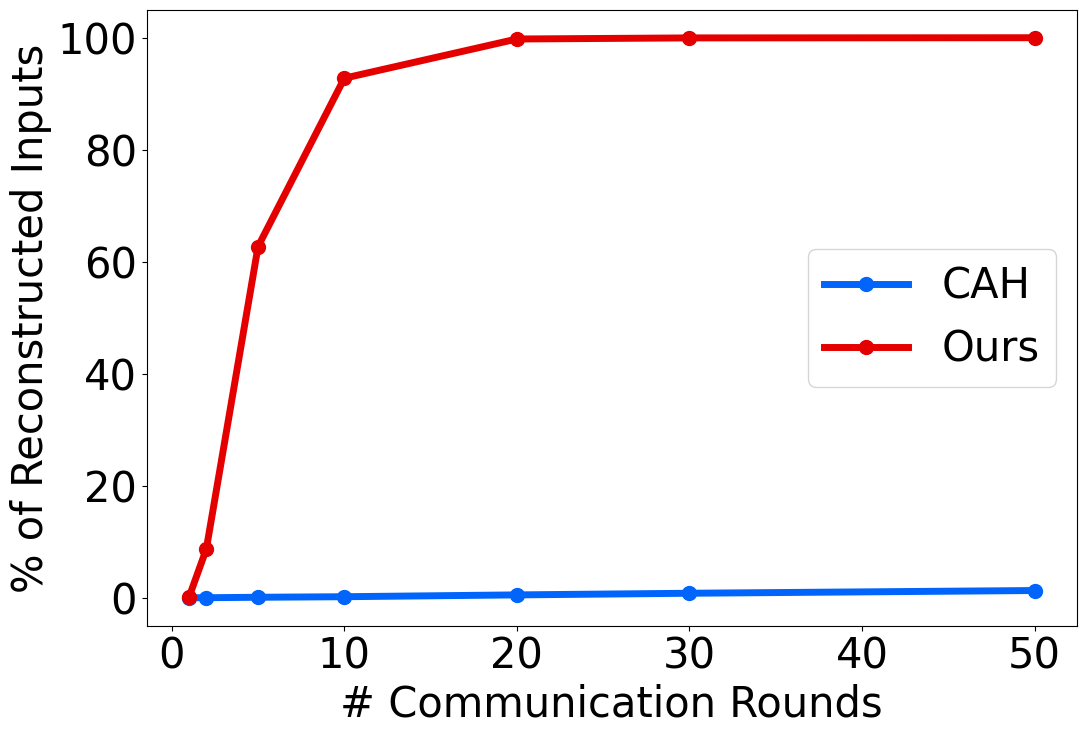}
    }
    \caption{a) Effect of batch size on CIFAR-10 after 20 communication rounds. b) Effect of the number of rounds on CIFAR-10 on a batch containing 2048 images.}

    % \vskip\baselineskip % Add vertical space between rows

\end{figure*}

\begin{figure}[ht]
    \centering
    \includegraphics[scale=0.25]{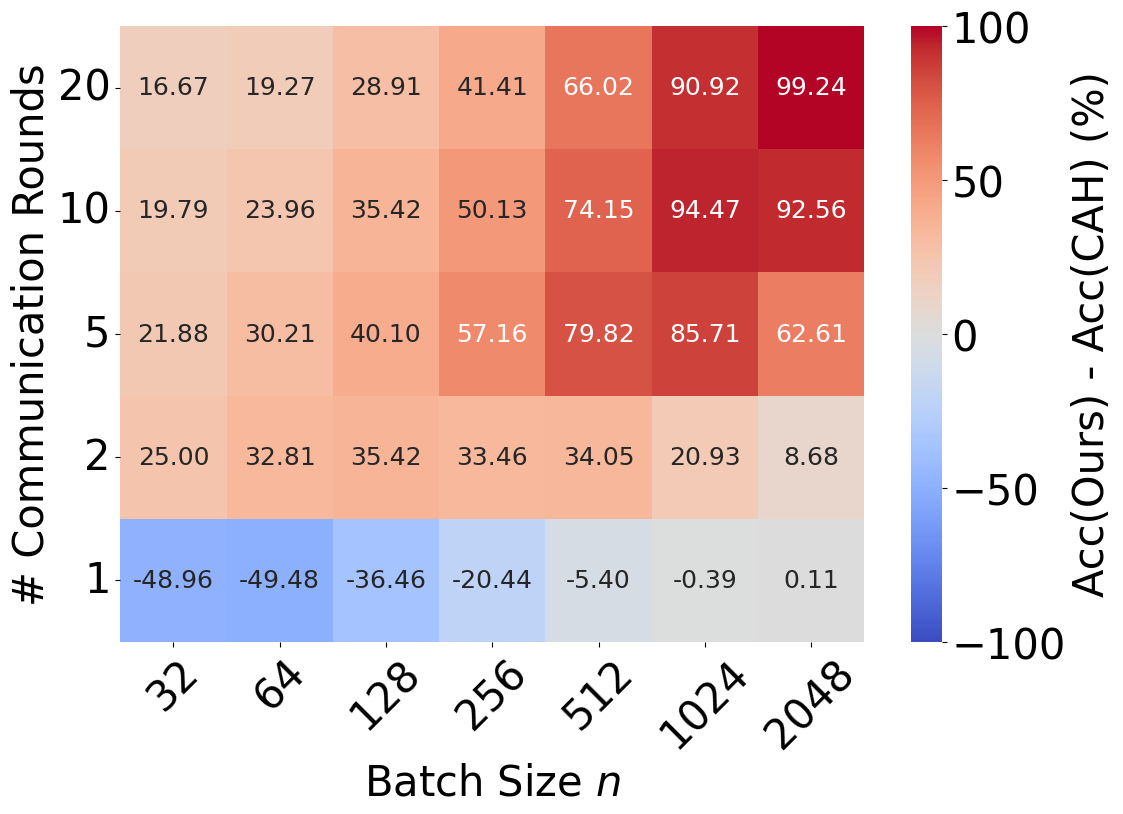}
    \caption{Percentage difference in reconstructed inputs on CIFAR-10. Red indicates cases where our attack achieves better reconstruction, while blue highlights instances where the baseline attack performs better.}
    \label{fig:heatmap_cifar}
\end{figure}

\paragraph{FC-NN} We conduct additional experiments on the CIFAR-10 dataset \cite{cifar10}, evaluating our attack on the same two-layers neural network described in the main paper. For these experiments, our malicious model parameters are initialized according to the setup used in the ImageNet experiments, described in App.~\ref{app:model_init}, while for CAH, we set $s=0.95$, according to the setup proposed in \citep{curious}. We perform the attack for batch size in \{32, 64, 128, 256, 512, 1024, 2048\} and evaluate after \{1, 2, 5, 10, 20\} rounds. Figure \ref{fig:batch_eff_cifar} illustrates the percentage of reconstructed images after 20 attack rounds. Our attack achieves perfect reconstruction for 2048 data points, while the baseline recovers less than 5\% of the samples. Figure \ref{fig:rounds_eff_cifar} shows the effect of the number communication rounds for $n=2048$. Our attack reaches perfect reconstruction after 20 rounds, while CAH's performance remains below 5\% of accuracy. Finally, Figure~\ref{fig:heatmap_cifar} provides a comprehensive comparison of the accuracy difference between our attack and the baseline across various batch sizes and communication rounds. CAH initially outperforms our method on image data after a single communication round. However, our attack surpasses the baseline after just two rounds,  achieving a remarkable 99\% accuracy advantage for $n=2048$ after 20 rounds and confirming the same trends observed on ImageNet and HARUS.

% \begin{figure}[htb]
%     \centering
%     \includegraphics[scale=0.15]{figures/cifar/rounds_effect.png}
%     \caption{Effect of communication on attack on CIFAR-10. Results are obtained considering a batch of 2048 images.}
%     \label{fig:com_cifar}
% \end{figure}

\paragraph{CNN}
To assess performance on a convolutional architecture, we test our attack on a VGG-style network. This network comprised three convolutional layers followed by two dense layers, consistent with the model architecture detailed in \cite[Table~9]{curious}. The experiments are executed in single precision. Although we observe slight numerical instability for the largest batch sizes that prevents us from fully recover the batch, we successfully reconstruct over 98\% of a 512-data point batch, demonstrating the attack's effectiveness even in a CNN setting.

\begin{table}[htb]
    \caption{Effect of of batch size and number of communication rounds. Each value indicates the percentage of data exactly recovered by attacks on CIFAR-10 dataset.}
    \label{tab:cnn_cifar}
    \centering
    \begin{tabular}{|c|c|c|c|c|c|c|c|}
    \hline
    \textbf{\# Rounds} & \multicolumn{6}{|c|}{\textbf{Batch size}}\\
    \hline
    & & 32 & 64 & 128 & 256 & 512 \\
     \hline
\multirow{2}{*}{1} & Ours &32.29\scriptsize{$\pm$11.83} &13.54\scriptsize{$\pm$3.93} &5.47\scriptsize{$\pm$1.56} &2.86\scriptsize{$\pm$1.13} &0.65\scriptsize{$\pm$0.30}  \\
                   & CAH  &66.67\scriptsize{$\pm$6.51} &56.77\scriptsize{$\pm$5.02} &45.31\scriptsize{$\pm$5.12} &23.57\scriptsize{$\pm$0.45} &6.71\scriptsize{$\pm$1.57}  \\
                   \hline
\multirow{2}{*}{2} & Ours &95.83\scriptsize{$\pm$7.22} &91.15\scriptsize{$\pm$2.39} &85.42\scriptsize{$\pm$1.80} &69.27\scriptsize{$\pm$3.13} &46.88\scriptsize{$\pm$2.82}  \\
                   & CAH  &70.83\scriptsize{$\pm$4.77} &64.58\scriptsize{$\pm$3.61} &53.12\scriptsize{$\pm$4.88} &34.11\scriptsize{$\pm$3.03} &11.00\scriptsize{$\pm$2.14}  \\
                   \hline

\multirow{2}{*}{5} & Ours &100.00\scriptsize{$\pm$0.00} &100.00\scriptsize{$\pm$0.00} &100.00\scriptsize{$\pm$0.00} &98.31\scriptsize{$\pm$2.29} &98.50\scriptsize{$\pm$0.98}  \\
                   & CAH  &80.21\scriptsize{$\pm$1.80} &70.83\scriptsize{$\pm$2.39} &60.68\scriptsize{$\pm$3.25} &44.40\scriptsize{$\pm$3.61} &20.38\scriptsize{$\pm$2.36}  \\
                   \hline

\multirow{2}{*}{10} & Ours &100.00\scriptsize{$\pm$0.00} &100.00\scriptsize{$\pm$0.00} &100.00\scriptsize{$\pm$0.00} &98.31\scriptsize{$\pm$2.29} &98.50\scriptsize{$\pm$0.98} \\
                   & CAH  &83.33\scriptsize{$\pm$1.80} &75.00\scriptsize{$\pm$2.71} &65.89\scriptsize{$\pm$2.26} &51.82\scriptsize{$\pm$0.81} &27.34\scriptsize{$\pm$2.49} \\
                   \hline

\hline
\multirow{2}{*}{20} & Ours &100.00\scriptsize{$\pm$0.00} &100.00\scriptsize{$\pm$0.00} &100.00\scriptsize{$\pm$0.00} &98.31\scriptsize{$\pm$2.29} &98.50\scriptsize{$\pm$0.98} \\
                   & CAH  &87.50\scriptsize{$\pm$3.13} &78.65\scriptsize{$\pm$2.39} &71.09\scriptsize{$\pm$2.07} &57.42\scriptsize{$\pm$1.35} &34.77\scriptsize{$\pm$2.17} \\
                   \hline

    \end{tabular}
\end{table}

Table~\ref{tab:cnn_cifar} compares our attack and the baseline on CIFAR-10. While the baseline effectively recovers small batches, its accuracy degrades significantly with larger batch sizes. In contrast, our attack consistently recovers over 98\% of the batch, even at larger sizes. Small reconstruction errors are likely due to numerical precision issues.

\paragraph{Multiple Local Steps}\label{par:local_steps}
In Sec~\ref{sec:exp_results} we assumed updates based on full-batch gradient computations. Here, we extend our attack to the setting in which clients perform $K > 1$ local gradient steps. As discussed in Sec.~\ref{sec:conclusions}, performing multiple local updates modifies the hyperplanes directions and positions, since $\Win$ and $\bin$ are updated at each step.  This complicates the precise isolation of individual data points in the local dataset of the attacked client.
To mitigate this effect, we propose a strategy that limits the influence of local updates. In particular, we aim to keep the gradients $\frac{\partial \loss}{\partial \Win}$ and $\frac{\partial \loss}{\partial \bin}$ small compared to the magnitude of $\Win$ and $\bin$, thereby ensuring that parameter updates remain minimal over multiple steps.
We consider a two-layer fully connected neural network, as described in the main paper. From \eqref{eq:control_grad} we observe that we can effectively limit the magnitude of the gradients by carefully designing the output weights. Specifically, by assigning a large value $\beta$ to half of the elements in $\Wout_i$ and a slightly smaller value $\beta - \eta$ to the other half (with $\eta$ being a small positive constant), the resulting gradients with respect to the bias $\bini$ and the input weights $\Win_i$ remain small—approximately bounded by $\eta$ and $\eta \x_j$, respectively. In our implementation, we use a vector $\mathbf{v} \in \mathbb{R}^C$, where the first half of its components are set to $10^5$, and the second half to $10^5 - \eta$ with $\eta = 10^{-2}$. The vector is then repeated across all the columns in $\Wout$. To avoid gradient cancellation effects (i.e., $\frac{\partial \mathcal{L}}{\partial \bin} = 0$) in case the training batch contains an equal number of samples from each half of the classes, we add small Gaussian noise $\mathcal{N}(0, 10^{-4})$ to each component of $\mathbf{v}$.

In parallel, we also aim to minimize updates to $\Wout$ itself across local steps. By application of the chain rule, we have $\frac{\partial \loss_j}{\partial \Wout}= \frac{\partial \loss_j}{\partial \z^{(2)}} \frac{\partial \z^{(2)}}{\partial \Wout}$,  where the first term is given by \eqref{eq:part_loss}, and the second term is $\z^{(1)} = \Win \x_j + \bin$. As the loss gradient $\frac{\partial \mathcal{L}j}{\partial \z^{(2)}}$ depends on classification prediction, and $\bin$ is modified to perform the binary search (Alg.\ref{algo:update_hp}), we can only directly control $\Win$.  Assuming that adjacent pixels in images have similar values, we set weights with alternating signs to drive the dot product $\Win_i \x_j$ to be close to zero. Specifically, we sample the even elements of $\Win_{i}$ from $U(1,2)$ and for odd ones as $\Win_{i} \sim U(-2,-1)$.

Due to the presence of local updates, it is no longer sufficient to determine whether two consecutive hyperplanes isolate a point by directly comparing their corresponding observations for exact equality. Instead, we assess whether the observations are sufficiently similar by projecting the observation associated with the larger bias value onto the other, and computing the Euclidean norm of the resulting projection vector. If this value is smaller than $10^{-4}$, we conclude that the corresponding hyperplanes do not separate any image.

 In our experiment, we assume that the target client holds $|S| = 1024$ images—here, $|S|$ denotes the dataset size of the client, which differs from the training batch size $n$. We set the number of local steps to $K = |S|/n$, with $n=\{2^3, 2^4, ..., 2^{10}\}$, and use a learning rate of $10^{-4}$, a standard choice for training on CIFAR10 to ensure convergence.

\begin{figure}[h]
    \includegraphics[width=0.5\linewidth]{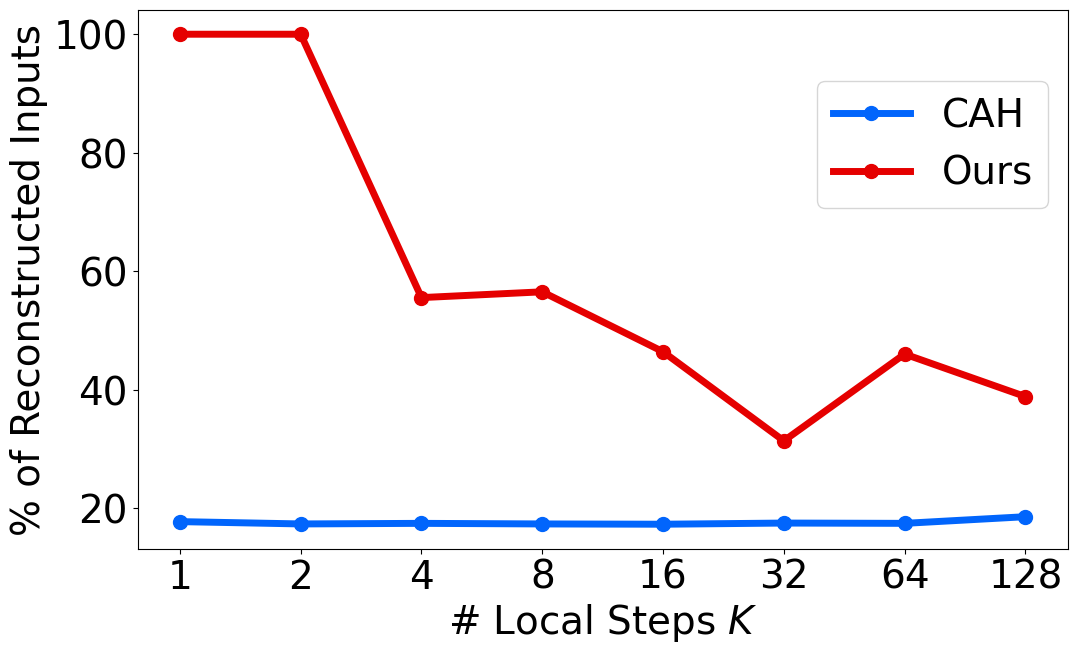}
    \centering
    \caption{ The percentage of perfectly reconstructed inputs  for a two-layer FC NN with $N=1000$ neurons in the first layer. The reconstruction is evaluated after $50$ communication rounds. } 
    \label{fig:minibatch}
\end{figure}

 The  results in Figure~\ref{fig:minibatch} demonstrate that increasing the number of local steps makes the attack more challenging; however, our method is still able to recover up to 35\% of the images even with local steps. In contrast, the baseline attack (CAH) fails to recover more than 20\% of the images under the same setting. It is also worth noting that in our setup we considered more local steps $K$ and larger dataset size than what considered in the baseline \cite{curious},as their original setting  involves $K=5$ local steps and a smaller dataset size ($|S| = 200$).

\paragraph{Robustness to Noise} To assess the robustness of our attack, we conduct experiments in the presence of noise, which may arise from imperfections in real-world communication channels or from defense mechanisms implemented by the client to defend against privacy attacks, such as local differential privacy\citep{dp_dwork}.

To improve robustness, the server can design model weights similarly to the strategy proposed to adapt to multiple local steps (check App.~\ref{par:local_steps}), but with a different objective. In that case, the server modifies the classification weights $\Wout$ to get a small value for $\frac{\partial \loss}{\partial \Win}$ and $\frac{\partial \loss}{\partial \bin}$ , while here the objective is the opposite: the attacker needs to obtain gradients of large magnitude, so that noise can be filtered out and will not have an impact on the reconstruction quality.

Hence, we define a vector $\mathbf{v} \in \mathbb{R}^C$ by setting the first half of its components to $10^5$ and the second half to $10^5 - \eta$, where $\eta = 10^3$. This vector is then replicated across all columns of $\Wout$.

\begin{figure}[h]
    \includegraphics[width=0.6\linewidth]{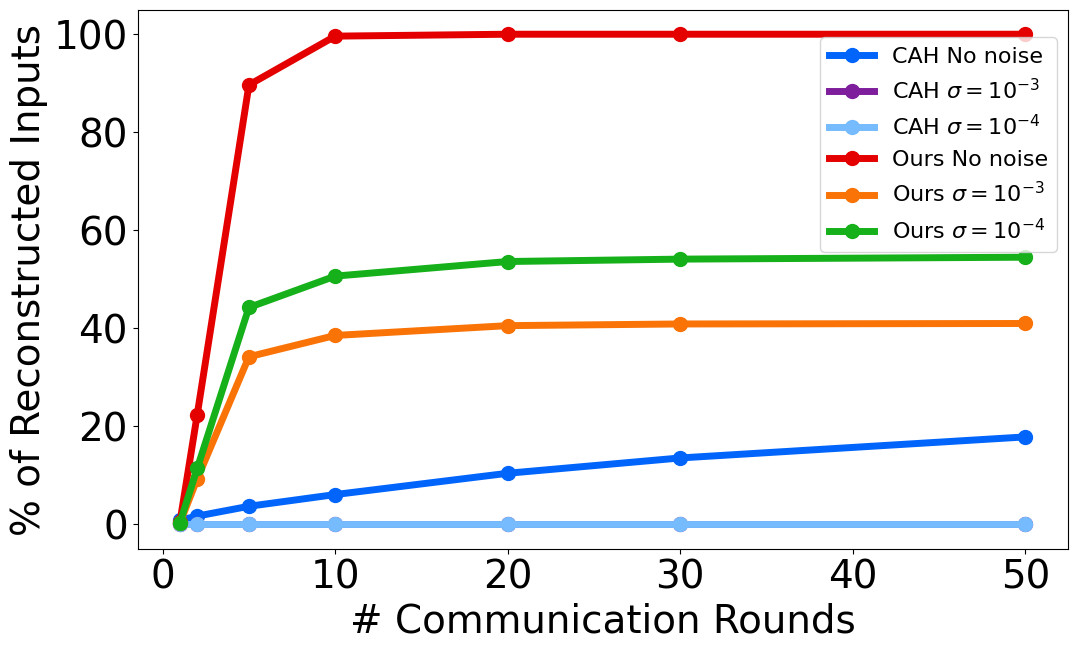}
    \centering
    \caption{ The percentage of perfectly reconstructed inputs on for two-layer fully connected neural network with $N=1000$ neurons in the first layer, under varying numbers of attack rounds and different noise variances $\sigma$.} 
    \label{fig:noise_effect}
\end{figure}

Figure \ref{fig:noise_effect} illustrates the attack performance across different numbers of rounds and varying Gaussian noise variances. The client holds 1024 images, trains using FedSGD with full-batch updates, and uses a learning rate of $10^{-4}$. As expected, the attack accuracy decreases with increasing noise variance; however, our method remains substantially more effective than the baseline (CAH). While our attack performance degrades gracefully as more noise is added on the client side, CAH is unable to reconstruct any sample for both values of $\sigma$ considered (the two corresponding curves overlap). As a consequence, our attack consistently outperforms the baseline, recovering almost 40\% more images.

\subsection{HARUS}
We investigate the effect of numerical precision by performing additional HARUS experiments in single precision. Table~\ref{tab:harus_sp} shows the results of both our attack and the baseline. The performance of our attack are comparable with the ones presented in the main paper, showing that the attack's success does not depend on double precision. However, also in this case precision errors can prevent us from recovering the full batch, as illustrated by the case of 1024 samples where extending the attack from 20 to 50 rounds does not improve the reconstruction rate.

\begin{table}[h]
    \centering
    \caption{Impact of batch size and number of communication rounds. Each value indicates the percentage of data exactly recovered in attacks on HARUS dataset.}
    \label{tab:harus_sp}
    \begin{tabular}{|c|c|c|c|c|c|c|c|c|c|}
    \hline
    \textbf{\# Rounds} & \multicolumn{9}{|c|}{\textbf{Batch size}}\\
    \hline
        & & 32 & 64 & 128 & 256 & 512 & 1024 & 2048 & 4096 \\
\hline
\multirow{2}{*}{1} & Ours & 33.33\scriptsize{$\pm$22.17} & 13.54\scriptsize{$\pm$6.31} & 5.47\scriptsize{$\pm$2.34} & 1.69\scriptsize{$\pm$0.81} & 1.24\scriptsize{$\pm$0.96} & 0.72\scriptsize{$\pm$0.91} & 0.29\scriptsize{$\pm$0.27} & 0.15\scriptsize{$\pm$0.06} \\
& CAH & 51.04\scriptsize{$\pm$4.77} & 23.96\scriptsize{$\pm$1.80} & 10.16\scriptsize{$\pm$0.78} & 4.30\scriptsize{$\pm$1.41} & 1.69\scriptsize{$\pm$0.41} & 0.55\scriptsize{$\pm$0.06} & 0.20\scriptsize{$\pm$0.08} & 0.07\scriptsize{$\pm$0.04} \\
\hline
\multirow{2}{*}{2} & Ours & 97.92\scriptsize{$\pm$3.61} & 95.83\scriptsize{$\pm$4.77} & 86.20\scriptsize{$\pm$4.71} & 71.09\scriptsize{$\pm$4.75} & 48.96\scriptsize{$\pm$6.88} & 25.72\scriptsize{$\pm$6.31} & 9.47\scriptsize{$\pm$1.50} & 2.90\scriptsize{$\pm$0.60} \\
& CAH & 73.96\scriptsize{$\pm$9.55} & 39.06\scriptsize{$\pm$3.13} & 17.19\scriptsize{$\pm$3.91} & 7.29\scriptsize{$\pm$2.35} & 3.45\scriptsize{$\pm$0.92} & 1.11\scriptsize{$\pm$0.30} & 0.42\scriptsize{$\pm$0.15} & 0.12\scriptsize{$\pm$0.02} \\
\hline
\multirow{2}{*}{5} & Ours & 100.0\scriptsize{$\pm$0.00} & 100.0\scriptsize{$\pm$0.00} & 100.0\scriptsize{$\pm$0.00} & 100.0\scriptsize{$\pm$0.00} & 97.01\scriptsize{$\pm$0.60} & 87.73\scriptsize{$\pm$2.99} & 62.50\scriptsize{$\pm$3.09} & 32.10\scriptsize{$\pm$2.12} \\
& CAH & 93.75\scriptsize{$\pm$3.13} & 69.27\scriptsize{$\pm$5.02} & 31.51\scriptsize{$\pm$0.45} & 14.58\scriptsize{$\pm$0.81} & 6.77\scriptsize{$\pm$0.11} & 2.83\scriptsize{$\pm$0.35} & 1.14\scriptsize{$\pm$0.12} & 0.33\scriptsize{$\pm$0.06} \\
\hline

\multirow{2}{*}{10} & Ours & 100.0\scriptsize{$\pm$0.00} & 100.0\scriptsize{$\pm$0.00} & 100.0\scriptsize{$\pm$0.00} & 100.0\scriptsize{$\pm$0.00} & 99.87\scriptsize{$\pm$0.23} & 99.41\scriptsize{$\pm$0.34} & 92.95\scriptsize{$\pm$0.79} & 70.61\scriptsize{$\pm$2.26} \\
& CAH & 96.88\scriptsize{$\pm$0.00} & 82.81\scriptsize{$\pm$2.71} & 45.31\scriptsize{$\pm$1.56} & 22.66\scriptsize{$\pm$2.17} & 10.81\scriptsize{$\pm$1.08} & 4.13\scriptsize{$\pm$0.50} & 1.71\scriptsize{$\pm$0.13} & 0.62\scriptsize{$\pm$0.01} \\
\hline

\multirow{2}{*}{20} & Ours &100.00\scriptsize{$\pm$0.00} &100.00\scriptsize{$\pm$0.00} &100.00\scriptsize{$\pm$0.00} &100.00\scriptsize{$\pm$0.00} &99.87\scriptsize{$\pm$0.23} &99.93\scriptsize{$\pm$0.11} &99.87\scriptsize{$\pm$0.06} &94.48\scriptsize{$\pm$0.59} \\
                    & CAH  &100.00\scriptsize{$\pm$0.00} &94.27\scriptsize{$\pm$0.90} &60.94\scriptsize{$\pm$2.07} &32.81\scriptsize{$\pm$1.95} &16.15\scriptsize{$\pm$2.29} &6.87\scriptsize{$\pm$1.41} &2.99\scriptsize{$\pm$0.31} &1.04\scriptsize{$\pm$0.05} \\
                    \hline
\multirow{2}{*}{30} & Ours &100.00\scriptsize{$\pm$0.00} &100.00\scriptsize{$\pm$0.00} &100.00\scriptsize{$\pm$0.00} &100.00\scriptsize{$\pm$0.00} &99.87\scriptsize{$\pm$0.23} &99.93\scriptsize{$\pm$0.11} &99.93\scriptsize{$\pm$0.06} &99.04\scriptsize{$\pm$0.14} \\
                    & CAH  &100.00\scriptsize{$\pm$0.00} &96.88\scriptsize{$\pm$1.56} &69.79\scriptsize{$\pm$0.45} &39.84\scriptsize{$\pm$1.41} &20.51\scriptsize{$\pm$1.19} &9.15\scriptsize{$\pm$1.48} &4.04\scriptsize{$\pm$0.10} &1.38\scriptsize{$\pm$0.08} \\
                    \hline

\multirow{2}{*}{50} & Ours &100.00\scriptsize{$\pm$0.00} &100.00\scriptsize{$\pm$0.00} &100.00\scriptsize{$\pm$0.00} &100.00\scriptsize{$\pm$0.00} &99.87\scriptsize{$\pm$0.23} &99.93\scriptsize{$\pm$0.11} &99.93\scriptsize{$\pm$0.05} &99.90\scriptsize{$\pm$0.13} \\
                    & CAH  &100.00\scriptsize{$\pm$0.00} &97.92\scriptsize{$\pm$2.39} &79.43\scriptsize{$\pm$3.16} &50.78\scriptsize{$\pm$2.03} &28.52\scriptsize{$\pm$1.19} &13.41\scriptsize{$\pm$1.94} &6.01\scriptsize{$\pm$0.42} &2.23\scriptsize{$\pm$0.10} \\
                    \hline

    \end{tabular}
\end{table}

\subsection{ImageNet}

\begin{figure}[t]
    \centering
    \includegraphics[scale=0.25]{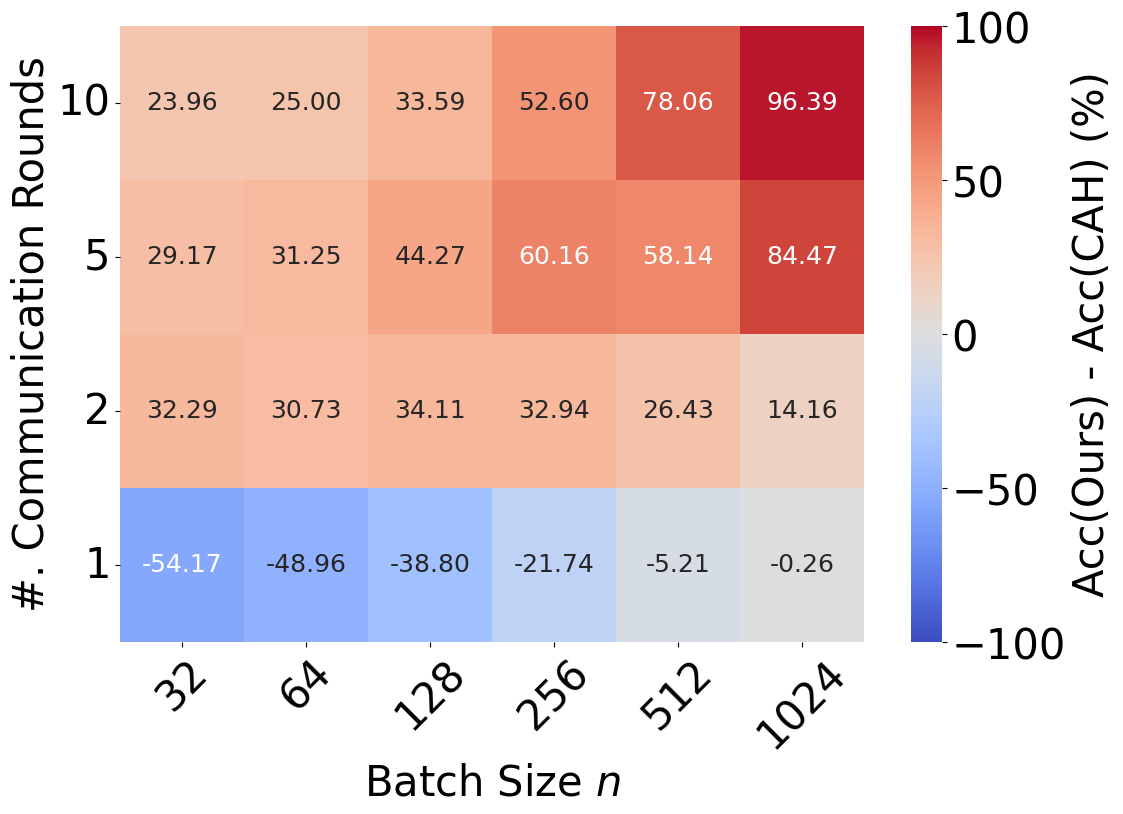}
    \caption{Percentage difference in reconstructed inputs on ImageNet. Red indicates cases where our attack achieves better reconstruction, while blue highlights instances where the baseline attack performs better.}
    \label{fig:heatmap_imagenet}
\end{figure}

Figure \ref{fig:heatmap_imagenet} evaluates the difference in reconstruction accuracy between our proposed attack and the baseline across varying batch sizes and communication rounds on ImageNet, confirming our attack's superior performance over CAH in all multi-round scenarios. The prevalence of red, especially in the upper rows and to the right, clearly demonstrates that our attack generally outperforms the baseline in most settings. This highlights the effectiveness of our approach in leveraging multiple communication rounds and larger batch sizes to improve reconstruction accuracy. The blue cells at the bottom of the heatmap, corresponding to the first communication round, suggest that CAH might have an advantage in single-round attacks, particularly for smaller batch sizes, where the baseline benefits from the multiple-directions initialization. Overall, results confirm the same trends as those on HARUS in the main paper.

\paragraph{Non-IID setting}
We evaluated the performance of our attack under both IID and non-IID data distributions. Table~\ref{tab:effect_data_dist} reports results on the ImageNet dataset using a single full-batch update, across different batch sizes and numbers of attack rounds.

In the IID setting, each client samples images uniformly at random from all 1,000 available classes. To simulate a non-IID scenario, we assign each client $i$ a distinct subset of classes $\mathcal{C}_i$ such that $\mathcal{C}_i\cap \mathcal{C}_j = \emptyset$, for any $i\not=j$. The result is that the local datasets of two clients do not share any common label. Each local dataset $S_i$ is constructed by sequentially sampling 50 images per class from $\mathcal{C}_i$ until the desired dataset size is reached.
Our results indicate that the attack remains highly effective in both settings. In particular, the attacker successfully reconstructs over 99\% of the dataset within 5, 10, and 20 attack rounds for batch sizes of 256, 512, and 1024, respectively. While performance in the non-IID setting is occasionally slightly worse than in the IID case, this difference is marginal. One possible explanation lies in the spatial distribution of the data: the expected distance between two images randomly sampled from a subset of classes is smaller than the expected distance between two images sampled from the full set of classes. As a result, from a geometrical perspective, it becomes more challenging to position hyperplanes that isolate a single image, as $\Delta$ in \eqref{eq:final_rounds_bound} will be smaller than in the IID case.

\begin{table*}[htb]
    \caption{Effect of client data distribution. Each value indicates the percentage of data exactly recovered in attacks on ImageNet dataset.}
    \label{tab:effect_data_dist}
    \medskip
    \centering
    \footnotesize
    \begin{tabular}{|c|c|c|c|c|c|c|c|}
    \hline
    \textbf{\# Rounds} & \multicolumn{7}{|c|}{\textbf{Batch size}}\\
    \hline
    & & 32 & 64 & 128 & 256 & 512& 1024 \\
    \hline
    \multirow{2}{*}{2} 
    & non-IID & 93.75\scriptsize{$\pm$6.25} & 90.63\scriptsize{$\pm$1.56} & 82.29\scriptsize{$\pm$1.97} & 63.80\scriptsize{$\pm$3.16} & 37.43\scriptsize{$\pm$2.85} & 16.34\scriptsize{$\pm$1.64} \\
    & IID     & 96.88\scriptsize{$\pm$3.13} & 93.75\scriptsize{$\pm$4.13} & 82.81\scriptsize{$\pm$5.90} & 63.80\scriptsize{$\pm$4.82} & 36.00\scriptsize{$\pm$2.67} & 14.78\scriptsize{$\pm$0.88} \\
    \hline
    \multirow{2}{*}{5} 
    & non-IID & 100.00\scriptsize{$\pm$0.00} & 100.00\scriptsize{$\pm$0.00} & 100.00\scriptsize{$\pm$0.00} & 99.74\scriptsize{$\pm$0.45} & 95.38\scriptsize{$\pm$1.30} & 81.64\scriptsize{$\pm$2.71} \\
    & IID     & 100.00\scriptsize{$\pm$0.00} & 100.00\scriptsize{$\pm$0.00} & 100.00\scriptsize{$\pm$0.00} & 99.74\scriptsize{$\pm$0.45} & 97.27\scriptsize{$\pm$1.74} & 86.10\scriptsize{$\pm$0.73} \\
    \hline
    \multirow{2}{*}{10} 
    & non-IID & 100.00\scriptsize{$\pm$0.00} & 100.00\scriptsize{$\pm$0.00} & 100.00\scriptsize{$\pm$0.00} & 100.00\scriptsize{$\pm$0.00} & 99.48\scriptsize{$\pm$0.23} & 94.95\scriptsize{$\pm$0.45} \\
    & IID     & 100.00\scriptsize{$\pm$0.00} & 100.00\scriptsize{$\pm$0.00} & 100.00\scriptsize{$\pm$0.00} & 100.00\scriptsize{$\pm$0.00} & 100.00\scriptsize{$\pm$0.00} & 99.71\scriptsize{$\pm$0.17} \\
    \hline
    \multirow{2}{*}{20} 
    & non-IID & 100.00\scriptsize{$\pm$0.00} & 100.00\scriptsize{$\pm$0.00} & 100.00\scriptsize{$\pm$0.00} & 100.00\scriptsize{$\pm$0.00} & 99.87\scriptsize{$\pm$0.23} & 99.32\scriptsize{$\pm$0.10} \\
    & IID     & 100.00\scriptsize{$\pm$0.00} & 100.00\scriptsize{$\pm$0.00} & 100.00\scriptsize{$\pm$0.00} & 100.00\scriptsize{$\pm$0.00} & 100.00\scriptsize{$\pm$0.00} & 99.84\scriptsize{$\pm$0.06} \\
    \hline
    \end{tabular}
\end{table*}

\paragraph{Single precision} We also perform single-precision ImageNet experiments, see Table \ref{tab:imagenet}.  Consistently with the HARUS results, single precision does not significantly limit performance, and we achieve near-complete batch recovery in most settings. Minor reconstruction errors can be attributed to numerical errors.
\begin{table*}[htb]
    \caption{Effect of batch size and number of communication rounds. Each value indicates the percentage of data exactly recovered in attacks on ImageNet dataset.}
    \label{tab:imagenet}
    \medskip
    \centering
    \footnotesize
    \begin{tabular}{|c|c|c|c|c|c|c|c|c|}
    \hline
    \textbf{\# Rounds} & \multicolumn{7}{|c|}{\textbf{Batch size}}\\
    \hline
    & & 32 & 64 & 128 & 256 & 512& 1024 \\
     \hline
\multirow{2}{*}{1} & Ours &3.13\scriptsize{$\pm$0.00} &2.60\scriptsize{$\pm$0.90} &0.78\scriptsize{$\pm$0.78} &0.13\scriptsize{$\pm$0.23} &0.20\scriptsize{$\pm$0.00} &0.10\scriptsize{$\pm$0.00} \\
                   & CAH  &56.25\scriptsize{$\pm$8.27} &51.56\scriptsize{$\pm$0.00} &39.58\scriptsize{$\pm$1.80} &21.88\scriptsize{$\pm$3.13} &5.40\scriptsize{$\pm$2.28} &0.39\scriptsize{$\pm$0.35} \\
                   \hline
\multirow{2}{*}{2} & Ours &96.88\scriptsize{$\pm$3.13} &93.75\scriptsize{$\pm$4.13} &82.81\scriptsize{$\pm$5.90} &63.80\scriptsize{$\pm$4.82} &36.00\scriptsize{$\pm$2.67} &14.78\scriptsize{$\pm$0.88} \\
                   & CAH  &63.54\scriptsize{$\pm$4.77} &59.38\scriptsize{$\pm$3.13} &47.14\scriptsize{$\pm$0.90} &29.69\scriptsize{$\pm$3.34} &9.18\scriptsize{$\pm$3.53} &0.78\scriptsize{$\pm$0.64} \\
                   \hline
\multirow{2}{*}{5} & Ours &100.00\scriptsize{$\pm$0.00} &100.00\scriptsize{$\pm$0.00} &100.00\scriptsize{$\pm$0.00} &99.74\scriptsize{$\pm$0.45} &97.27\scriptsize{$\pm$1.74} &86.10\scriptsize{$\pm$0.73} \\
                   & CAH  &70.83\scriptsize{$\pm$1.80} &68.75\scriptsize{$\pm$1.56} &55.73\scriptsize{$\pm$0.90} &39.58\scriptsize{$\pm$0.60} &15.95\scriptsize{$\pm$4.40} &1.82\scriptsize{$\pm$0.64} \\
                   \hline
\multirow{2}{*}{10} & Ours &100.00\scriptsize{$\pm$0.00} &100.00\scriptsize{$\pm$0.00} &100.00\scriptsize{$\pm$0.00} &100.00\scriptsize{$\pm$0.00} &100.00\scriptsize{$\pm$0.00} &99.71\scriptsize{$\pm$0.17} \\
                   & CAH  &76.04\scriptsize{$\pm$4.77} &75.00\scriptsize{$\pm$3.13} &66.41\scriptsize{$\pm$3.41} &47.40\scriptsize{$\pm$0.60} &21.94\scriptsize{$\pm$5.38} &3.32\scriptsize{$\pm$1.28} \\
                   \hline

    \end{tabular}
\end{table*}

\begin{figure*}
    \centering
    \includegraphics[scale=0.25]{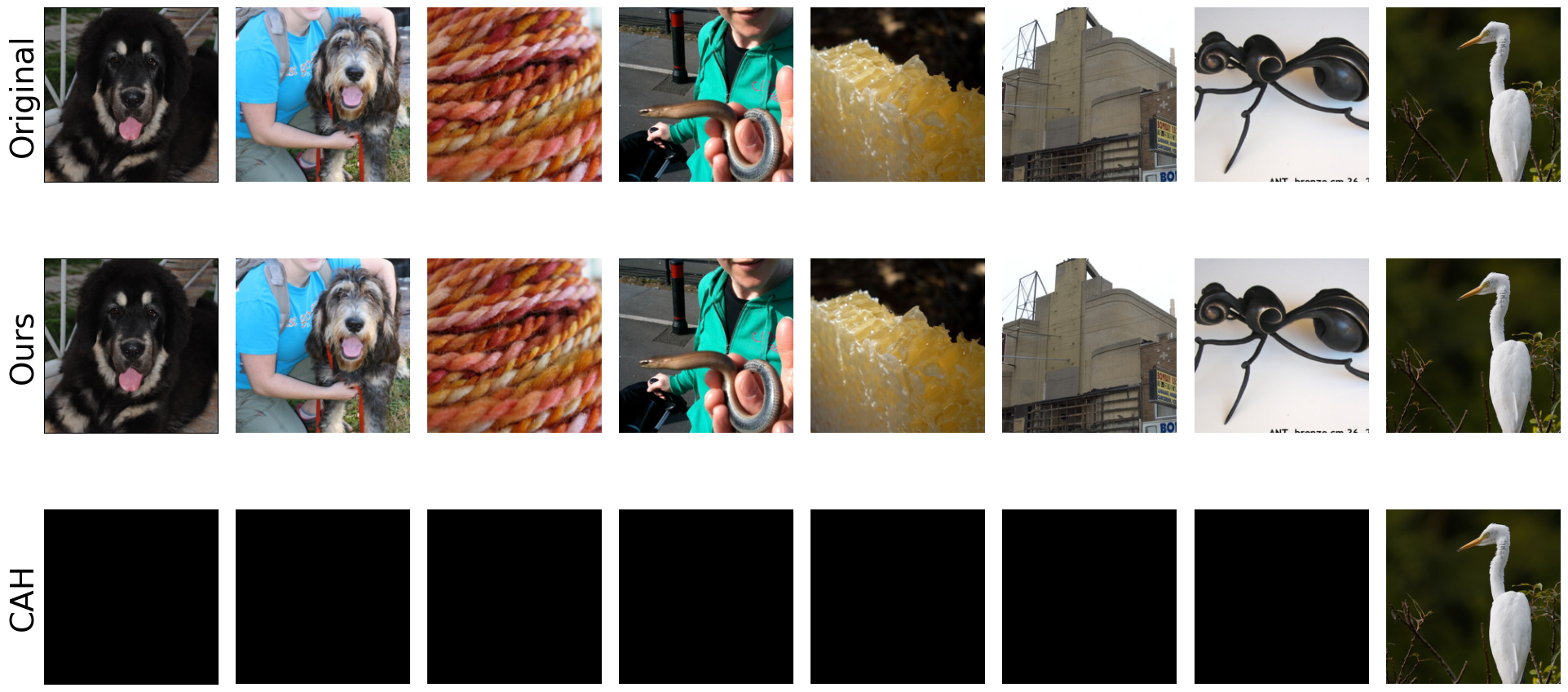}
     \caption{The recovered images from a batch containing 1024 samples from ImageNet. The first row shows the original images, the second row the recovery of our attack after 10 rounds, and the last row shows that the baseline is not able to recover most of the images. } \label{fig:images_examples1}

\end{figure*}

\begin{figure*}
    \centering
    \includegraphics[scale=0.3]{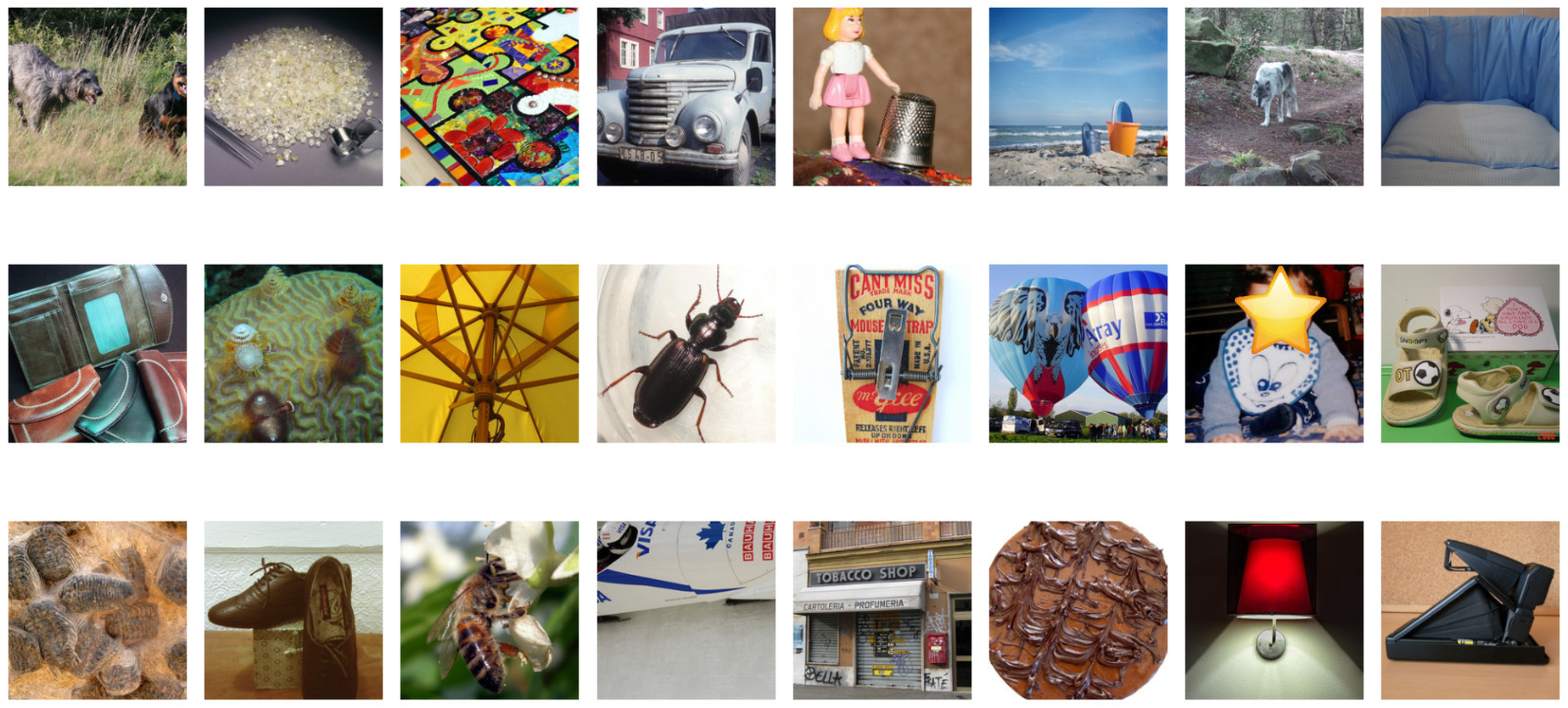}
     \caption{The recovered images from a batch containing 1024 samples from ImageNet.  }\label{fig:images_examples2}

\end{figure*}

% % \section{Results in Single Precision}\label{app:sing_prec_exp}
% % \begin{figure}
% %     \centering
% %     \includegraphics[width=0.5\linewidth]{batc_size_effect.png}
% %     \caption{}
% %     \label{fig:enter-label}
% % \end{figure}

\end{document}